\documentclass[11pt]{article}
\usepackage{tikz}
\usepackage[final]{acl}

\usepackage{times}
\usepackage{latexsym}
\usepackage{amsmath}
\usepackage{amssymb}
\usepackage{booktabs}
\usepackage{multirow} 
\usepackage{makecell}
\usepackage{pgfplots}
\usepgfplotslibrary{groupplots}
\pgfplotsset{compat=1.18}
\usepackage[T1]{fontenc}

\definecolor{oiBlue}{HTML}{0072B2}
\definecolor{oiOrange}{HTML}{D55E00}
\definecolor{oiGreen}{HTML}{009E73}
\definecolor{oiGray}{HTML}{666666}
\usepackage[utf8]{inputenc}

\usepackage{microtype}

\usepackage{inconsolata}

\usepackage{graphicx}
\usepackage{tcolorbox} % 必须引入这个包
\usepackage{listings}
\usepackage{amsmath}
\usepackage{algorithm}
\usepackage{algorithmic}
\usepackage{amssymb}

\title{Robust Code RL via Faulty-Code-Driven Test case Synthesis and Dense Reward Shaping}

\author{
 \textbf{Yiwen Zhang\textsuperscript{1,2}},
 \textbf{Xiaodong Yan\textsuperscript{2}},
 \textbf{Zhenyu Huang\textsuperscript{2}},
 \textbf{Deng Zhao\textsuperscript{2}},
 \textbf{Liang Jiang\textsuperscript{2}}, \\
 \textbf{Qing Cui\textsuperscript{2}},
 \textbf{Zujie Wen\textsuperscript{2}},
 \textbf{Zhiqiang Zhang\textsuperscript{2}},
 \textbf{Jun Zhou\textsuperscript{2}}\thanks{Corresponding authors}
\\
 \textsuperscript{1}Zhejiang University,
 \textsuperscript{2}Ant Group
\\
 \small{
   \textbf{Correspondence:} jun.zhoujun@antgroup.com
}
\\
}

\def \OURS{RobustTests}
\lstdefinestyle{NormalStyle}{
    basicstyle=\small\ttfamily,
    escapeinside={(*@}{@*)}, % 定义逃逸字符
    breaklines=true,      % 自动换行
    frame=single,         % 单线框
    backgroundcolor=\color{gray!10},
    rulecolor=\color{black},
    xleftmargin=0em,
    xrightmargin=0em
}

\lstdefinestyle{TinyStyle}{
    basicstyle=\scriptsize\ttfamily,
    escapeinside={(*@}{@*)}, % 定义逃逸字符
    breaklines=true,      % 自动换行
    frame=single,         % 单线框
    backgroundcolor=\color{gray!10},
    rulecolor=\color{black},
    xleftmargin=0em,
    xrightmargin=0em
}

\begin{document}
\maketitle
\begin{abstract}

Reinforcement Learning from Verifiable Rewards (RLVR) is pivotal for enhancing LLM code generation, yet its efficacy is often hindered by insufficient test case coverage, leading to reward hacking and policy degradation. To address this, we propose ~\OURS{}, a framework featuring a faulty-code-driven test case synthesis strategy. By leveraging "near-correct" faulty codes, ~\OURS{} captures latent logical discrepancies and employs validator agents with behavioral feature clustering to filter invalid or redundant test cases. Additionally, a stepwise dense reward function based on pass rates is introduced to mitigate false negatives and enhance training robustness. Using this pipeline, we construct an augmented version of the CodeContests$^{+}$ dataset with superior diagnostic utility. Experimental results show that RL fine-tuning of Qwen3-32B via ~\OURS{} achieves a 3\% absolute gain on LiveCodeBench, demonstrating its effectiveness in advancing LLM code generation proficiency. Codes and data are available at \url{https://huggingface.co/datasets/sid6/RobustTests}.
\end{abstract}

% 你是一位科研论文审稿员，擅长写作高质量的英文科研论文。请你帮我准确且学术性地将以下中文翻译成英文，风格与英文科研论文保持一致。给定的论文题目是Robust Code RL via Faulty Code Driven Test case Generation and Dense Reward Shaping。对我给定的每个段落，请注意上下文专有名词翻译的一致性和上下文的连贯性。

% \#\# 规则：

% - 输入格式为 Markdown 格式，输出格式也必须保留原始 Markdown 格式

% - 以下是常见的相关术语词汇对应表（中文 -> English）：

% * 强化学习 -> reinforcement learning
% * 测试用例 -> test case
% * 错误代码 -> faulty code
% * 错误程序 -> faulty code
% * 缺陷代码 -> faulty code

% \#\# 策略：

% 分三步进行翻译工作，并打印每步的结果：

% 1. 根据中文内容直译成英文，保持原有格式，不要遗漏任何信息

% 2. 根据第一步直译的结果，指出其中存在的具体问题，要准确描述，不宜笼统的表示，也不需要增加原文不存在的内容或格式，包括不仅限于：

% - 不符合英文表达习惯，明确指出不符合的地方

% - 语句不通顺，指出位置，不需要给出修改意见，意译时修复

% - 晦涩难懂，模棱两可，不易理解，可以尝试给出解释

% 3. 根据第一步直译的结果和第二步指出的问题，重新进行意译，保证内容的原意的基础上，使其更易于理解，更符合英文科研论文的表达习惯，同时保持原有的格式不变
\section{Introduction}

\begin{figure}[htbp]
    \includegraphics[width=1\linewidth]{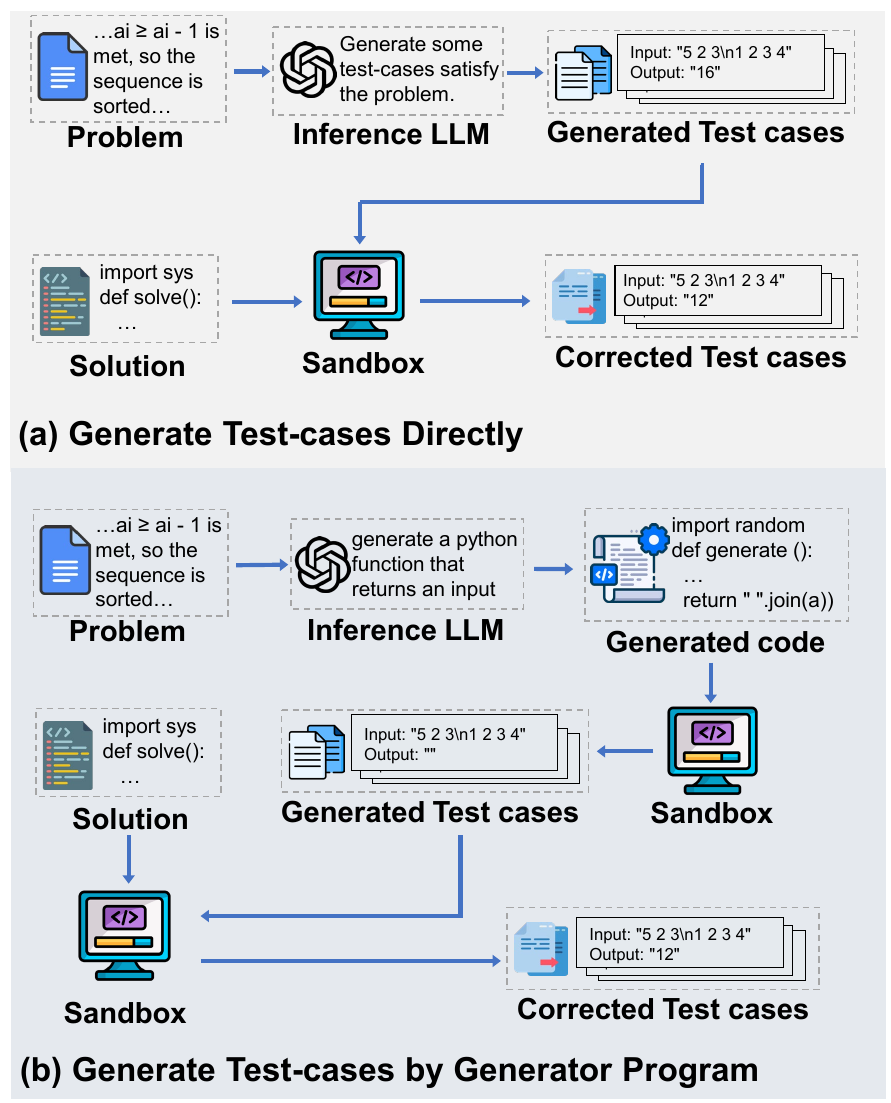}
    \caption{Illustration of current methods for test cases synthesis by LLM.(a) \textbf{Generate Test cases Directly}, instructing the LLM to directly synthesize a suite of test cases compliant with the problem specifications.(b) \textbf{Generate Test cases by Generator Program}, where LLM is first instructed to generate generator programs designed to automate the synthesis of test case inputs. }
    \label{fig:concept}
\end{figure}
% Substantial literature has demonstrated that post-training techniques centered on reinforcement learning (RL) exhibit significant promise in markedly enhancing the code generation capabilities of Large Language Models (LLMs)~\cite{el2025competitive}. Such research typically conceptualizes programming tasks within the Reinforcement Learning with Verifiable Rewards (RLVR) paradigm~\cite{jiang2026survey, hou2024large}. In this framework, models generate candidate solutions based on natural language instructions, with an automated verifier providing reward signals derived from execution results against test cases. However, the efficacy of RLVR is highly contingent upon the comprehensiveness of the test cases. Under the constraints of a limited test case scale, faulty code snippets may coincidentally satisfy a sparse suite of test cases, leading to false positives where the model erroneously classifies faulty code as correct.~\cite{le2022coderl}. In extreme scenarios, models may develop reward hacking strategies, such as hard-coding or brute-force result enumeration to circumvent complex logical reasoning~\cite{guo2025deepseek}. These behaviors severely contaminate RL training signals, triggering Policy Degeneracy and thereby stalling further capability gains. Consequently, the automated synthesis of high-quality, high-coverage test cases has become a pivotal strategy for refining RL feedback mechanisms and ensuring sustained performance elevation~\cite{ma2025dynamic, lin2025learning}.

Reinforcement Learning from Verifiable Rewards (RLVR) has emerged as a pivotal technique for enhancing the code generation capabilities of Large Language Models (LLMs)~\cite{el2025competitive, jiang2026survey,hou2024large}. However, its efficacy is fundamentally limited by the comprehensiveness of test cases. Insufficient coverage often causes false positives~\cite{le2022coderl}, where faulty code passes sparse test cases, leading to reward hacking and subsequent policy degradation\cite{guo2025deepseek}. Consequently, synthesizing high-coverage test cases is essential to refine RL feedback and ensure sustained performance gains~\cite{ma2025dynamic, lin2025learning}.

Nevertheless, the automated generation of high-quality test cases and their subsequent integration into a performance-boosting RLVR paradigm remain formidable challenges. Firstly, the lack of precise metrics to characterize test case quality obscures which data distributions are truly conducive to the RLVR process~\cite{liu2023your}. Secondly, the optimal methodology for integrating even high-quality test cases into RLVR frameworks remains under-explored~\cite{gunjal2025rubrics}. Although existing attempts leverage LLMs for direct test case augmentation (e.g., via few-shot prompting with problem descriptions)~\cite{zeng2025acecoder, xu2025kodcode, li2024large}, as shown in Figure~\ref{fig:concept}(a), the resulting outputs often fail to regard boundary condition coverage. Furthermore, test cases synthesized by this approach frequently exhibit hallucination instances that violate the underlying problem constraints. Adopting such erroneous verification signals as reward feedback in reinforcement learning induces significant reward bias, which misguides the optimization trajectory and constrains further enhancement of the model's proficiency.~\cite{kwa2024catastrophic}.

To enhance synthetic quality, subsequent efforts have transitioned to a generator program paradigm~\cite{he2025hardtests}, using various logical hypotheses to improve the coverage of test cases, as shown in Figure~\ref{fig:concept}(b). While this approach reduces the hallucination rate, it remains heavily reliant on manual annotation, which restricts its scalability. Building on this, \cite{wang2025codecontests+} introduced verification agents to automatically craft validator programs for constraint verification, further suppressing the hallucination rate. 
% However, even with automated validation, the Completeness of the verifiers lacks formal guarantees~\cite{olausson2023self}. As validator programs may not exhaustively cover all implicit constraints, the issue of residual hallucinations persists~\cite{liu2023your}. When employing traditional binary (0/1) sparse reward functions, evaluation feedback becomes exceedingly sensitive to the accuracy of the underlying test cases. If the test case suite harbors hallucinated test cases, even a correct solution generated by the model may be assigned a zero reward due to its failure to align with the erroneous labels. Such false negatives feedback generates misleading gradient signals that perturb the optimization trajectory of the model~\cite{ouyang2022training}. 
However, the completeness of automated validators is difficult to formally guarantee~\cite{olausson2023self}. Their intrinsic residual hallucinations, particularly under binary sparse rewards, frequently induce false negatives~\cite{liu2023your}. This results in correct code generating misleading gradient signals due to erroneous labeling~\cite{ouyang2022training}.
Consequently, within the context of RL reward signaling, simply improving data quality has hit a bottleneck. It is imperative to refine the learning mechanisms or introduce robust reward modeling~\cite{casper2023open} to mitigate the deleterious effects of residual hallucinations and ensure stable model evolution under noisy feedback.

To address these challenges, we propose \OURS, which at its core incorporates faulty-code-driven test case synthesis and a robust dense reward mechanism. During the synthesis stage, we first utilize LLMs to generate "near-correct" faulty programs refined via original test cases to guide the directed synthesis of test cases capable of triggering specific logical defects. Subsequently, we perform validity verification on these generated test cases to eliminate invalid ones. We further conduct clustering and filtering based on behavior feature vectors (i.e., the pass/fail status encodings of test cases over different faulty code snippets), which ensures that the test case suite can cover diverse error scenarios and alleviate false positives. During the training stage, to address the false negatives caused by unavoidable hallucinatory noise in synthetic test case suit, we introduce a stepwise dense reward function based on pass rates. This improves the robustness of the model to continuously learn from partially correct signals.

Within the \OURS{} framework, we augmented the test cases of the CodeContests$^{+}$ dataset to construct a code high-quality dataset, and evaluated its properties using Test case Space Polarization (TSP) metric. The results demonstrate that \OURS{} achieves a 7\% improvement in TSP relative to the original CodeContests$^{+}$, indicating its enhanced capacity to uncover a broader spectrum of failure modes. In our experiments, using problems of moderate difficulty from CodeContests$^{+}$ as the training set, RL fine-tuning of Qwen3-32B via \OURS{} yields an absolute 3\% performance gain on the LiveCodeBench benchmark compared to the baseline methods. These findings not only confirm the effectiveness of the \OURS{} framework in bolstering the code generation proficiency of LLMs but also establish a clear correlation between the TSP metric and model performance.

In summary, the main contributions of this study are as follows: 
\begin{itemize}
    \item We propose a novel approach for automated high-quality test case synthesis and RLVR reward modeling, utilizing faulty-code-driven synthesis and a robust dense reward mechanism to expand test case coverage and bolster resilience against synthetic noise during RL.
    \item We construct a code dataset with more diverse test cases, significantly strengthens diagnostic utility across various faulty code scenarios.
    \item An abosulte 3-percentage point performance gain on the LiveCodeBench benchmark when training Qwen3-32B compared to baselines, substantially advancing the code generation performance of LLMs.
\end{itemize}

\section{Related work}
\subsection{Test case synthesis Method}
%目前最构建测试用例最精确的方法是利用人工手动构建测试用例。很多代码评测基准（benchmark）的测试用例是采用此方法构建，如MBPP、HumanEval和 LiveCodeBench。但是，人工手动构建测试用例成本高且效率低下，因此仅适用于构建小规模评估集，用于大规模训练集构建时成本过高。因此，现在有一些基于LLM的构建方法，CodeContests通过对爬取的输入进行随机扰动生成额外测试用例，EvalPlus通过向大语言模型提供参考实现以合成种子输入，从而扩展 HumanEval 的测试用例，KodCode和 AceCoder做了进一步的拓展，不仅合成了测试用例，还利用此生成了编码问题和参考解决方案。HardTests方法对以上方法进行了改进，通过生成生成器程序来进行测试用例的生成，这种方法虽然增加了测试用例对不同错误代码的覆盖程度，但是高度依赖于人工对生成器程序的设定，仍然无法大规模的自动化。CodeContests$^{+}$在此基础上，通过自动编写验证器程序（Validator Programs）来校验输入约束，减少了大模型的幻觉问题。但是，如何生成能覆盖更多错误代码诊断的测试用例，仍然是一个具有挑战性的问题。
At present, the most accurate method for test case synthesis remains manual curation by human experts. This methodology underpins the test cases of numerous code evaluation benchmarks, including MBPP~\cite{austin2021program}, HumanEval~\cite{chen2021evaluating}, and LiveCodeBench~\cite{jain2024livecodebench}. However, it is prohibitively expensive and lacks scalability, rendering it suitable only for small-scale benchmark and impractical for the construction of massive training corpora. Consequently, several LLM-based automated methods have emerged. CodeContests$^{+}$~\cite{cai2026codecontests} generates supplementary test cases by applying stochastic perturbations to harvested inputs. EvalPlus~\cite{liu2023your} extends HumanEval by prompting LLMs to synthesize seed inputs guided by reference implementations. Furthermore, frameworks such as KodCode~\cite{xu2025kodcode} and AceCoder~\cite{zeng2025acecoder} have expanded this scope to include the joint synthesis of coding problems, test cases, and reference solutions. The HardTests~\cite{he2025hardtests} method improves upon these strategies by utilizing generator programs for test case synthesis. Building upon these, CodeContests$^{+}$~\cite{wang2025codecontests+} incorporates automated validator programs to verify input constraints, effectively mitigating LLM-induced hallucinations. Nevertheless, the challenge of synthesizing test cases that provide comprehensive diagnostic coverage for diverse faulty code remains an unresolved research problem.
\subsection{RL for Enhancing LLM's Code Ability}
% 强化学习（RL）已日益融入编程领域，利用代码可执行性提供客观反馈。早期研究，例如 CodeRL 和 AlphaCode，探索了使用编译器反馈或单元测试作为奖励信号，通过 PPO 等算法优化模型。DeepSeek-R1 在现有方法的基础上，引入了 GRPO 作为当前最先进的算法。然而，DeepSeek-R1 依赖于精心整理的公共数据集，但使用合成测试用例进行训练则需要不同的方法。由于合成测试用例不可避免地包含错误，因此必须设计一个鲁棒的奖励函数，以增强模型对测试用例不准确性的鲁棒性，并提供更具容错性的反馈信号。
Reinforcement learning (RL) has been increasingly integrated into the coding domain, leveraging code executability to provide objective feedback~\cite{jiang2026survey, shojaee2023execution, dou2024stepcoder}. Early studies, such as CodeRL~\cite{le2022coderl} and AlphaCode~\cite{li2022competition}, explored the use of compiler feedback or unit tests as reward signals to optimize models via algorithms like PPO~\cite{schulman2017proximal}. Building upon existing methodologies, DeepSeek-R1~\cite{guo2025deepseek} introduced the GRPO as the current state-of-the-art (SOTA) algorithm. However, while DeepSeek-R1 relies on curated public datasets, training with synthetic test cases necessitates a different approach. Since synthetic test cases inevitably harbor errors, it is imperative to design a robust reward function that enhances resilience toward test case inaccuracies and provides more tolerant feedback signals.

\section{Method}
\label{sec:method}
\begin{figure*}
    \centering
    \includegraphics[width=1\linewidth]{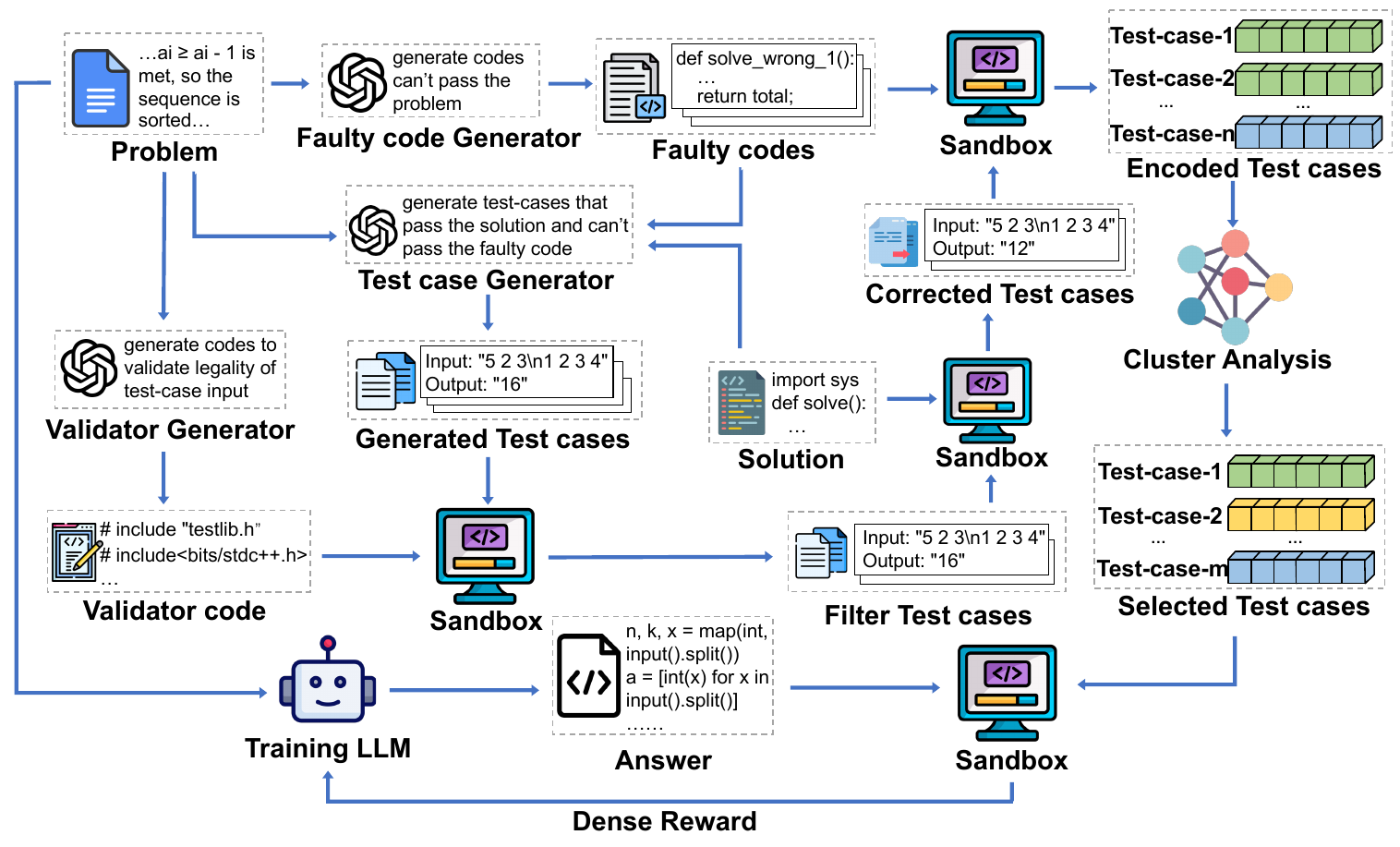}
    \caption{Overview of the proposed \OURS{} framework. The framework consists of three core components: (i) Automated Test case Generation in~\ref{sec:Synthesis}, aims to generate high-quality test cases inputs capable of effectively distinguishing correct code from faulty codes; (ii) Test case Validation and Selection in~\ref{sec:Selection}, ensures that the generated test cases are semantically sound and capable of diagnosing a broad spectrum of faulty code while simultaneously maximizing set parsimony; and (iii) Dense Reward Function Design in~\ref{sec: reward}, accounts for the potential fallibility of the synthesized test case suite.}
    \label{fig:overview}
\end{figure*}
% 为了解决利用基于可验证奖励的强化学习（RLVR）提升大语言模型（LLM）代码生成能力奖励偏差的问题，我们提出了\OURS，其核心包含错误代码驱动的测试用例构造与鲁棒性稠密奖励机制的设计，具体实现细节将在下文介绍。
To address the reward bias issues encountered when leveraging RLVR to enhance the code generation capabilities of LLMs, we introduce the \OURS{} framework, illustrated in Figure ~\ref{fig:overview}. 
% Detailed implementation of test case synthesis is provided in Sections ~\ref{sec:Synthesis} and~\ref{sec:Selection} and the robust dense reward mechanism is provided in Section~\ref{sec: reward}.
% \subsection{自动化测试用例合成}
% 测试用例合成的目标是生成能够有效区分正确代码与错误代码的高质量测试用例的输入。该过程由两个核心阶段组成：多样化错误代码池的构建，以及定向诱发失败的测试用例生成。
% \paragraph{错误代码池构建} 
% 为了给测试用例生成提供明确的目标，我们首先合成一组具有多样性的错误实现。针对每个问题 $P_i$，借鉴~\cite{yang2025swe}的思想，我们利用大语言模型（LLM）进行随机错误生成，通过采样获得广泛的潜在逻辑缺陷 $f \sim \text{LLM}(\cdot | P_i)$。为确保生成候选代码的非平凡性（Non-triviality），我们应用了严格的动态过滤机制。每个候选实现 $e$ 都会在基准测试集 $T_{\text{base}}$ 上运行，并产生一个二值执行向量 $V^f_i \in \{0,1\}^{|T_{\text{base}}|}$，其中每个维度代表相应测试用例的通过/失败状态。我们仅保留满足以下准则的错误代码：
% \begin{equation}
%     \phi(f) \triangleq 0 < \frac{\|V^f_i\|_1}{|T_{\text{base}}|} < 1
% \end{equation}
% 为了进一步优化代码库，我们通过对具有相同执行向量 $V^f_i$ 的错误代码执行精确去重（Exact Deduplication）。通过为每种唯一的失败模式仅选择一个代表性样本，得到候选池$F_{\text{base}}$。
% \paragraph{错误代码引导的测试用例生成} 
% 在获得合格的错误代码池 $F_{\text{base}}$ 后，我们的核心任务是合成能够探索“正确逻辑”与“具体实现陷阱”之间细微语义边界的测试用例。我们采用诱发失败提示策略，将批量的错误代码实现 作为负例输入LLM生成测试用例的输入。指令要求模型生成满足以下两个条件的符合约束的测试用例 $t_i$：(i) 严格遵守 $P_i$ 的输入规范，输出需要与输入对应；(ii) 能够在保持基准解正常运行的同时，诱发至少一个出现逻辑错误。我们将生成的所有测试用例纳入候选池$T_{\text{cand}}$中。

\subsection{Automated Test case Generation}
\label{sec:Synthesis}
Automated test case generation aims to generate high-quality test cases inputs capable of effectively distinguishing correct code from faulty codes. This process comprises two pivotal stages: the generation of a diverse pool of faulty code and the subsequent generation of directed, faulty-code-driven test cases. Representative samples of faulty code and generated test cases are detailed in Appendix~\ref{appendix: Faulty Codes and Generated Test cases}.
\paragraph{Faulty Code Generation} 
To establish concrete targets for test case generation, we first generate a diverse ensemble of faulty implementations. For each problem $P_i$, inspired by the approach in~\cite{yang2025swe}, we leverage an LLM to generate stochastic faulty code, producing a broad spectrum of potential logical defects $f \sim \text{LLM}(\cdot | P_i)$ via sampling. Prompts we used are shown in Appendix ~\ref{app: prompt1}. To ensure the non-triviality of the candidate implementations, we employ a rigorous dynamic filtering mechanism. Each candidate $e$ is executed against an original test case suite $T_{\text{base}}$, producing a binary execution vector $V^f_i \in \{0,1\}^{|T_{\text{base}}|}$, where each dimension denotes the pass/fail status of the corresponding test case. We retain only the faulty code that satisfies the following criterion:
\begin{equation}
    \phi(f) \triangleq 0 < \frac{\|V^f_i\|_1}{|T_{\text{base}}|} < 1
\end{equation}
The strategy filters out both fully correct and completely incorrect code while preserving only nearly correct faulty code. This ensures that the generated test cases are capable of capturing potential logical defects.

To further refine the faulty codes, we perform exact deduplication on implementations with the same execution vector $V^f_i$. By selecting a single representative sample for each unique failure mode, we construct the candidate pool $F_{\text{base}}$, thereby eliminating semantic redundancy and ensuring that the pool encompasses diverse logical discrepancies.
\paragraph{Faulty-Code-Driven Test case Generation} 
Given the qualified faulty code pool $F_{\text{base}}$, the primary objective is to generate test cases that probe the nuanced semantic boundaries between correct logic and specific implementation pitfalls. We adopt a failure-inducing prompting strategy~\cite{mu2024ddprompt}, where batches of faulty codes are provided as negative examples within the prompt. As shown in Appendix~\ref{app: prompt2}, The model is instructed to generate constraint-compliant test cases $t_i$ that satisfy two primary criteria: (i) strict adherence to the input specifications of $P_i$, ensuring that the generated outputs correspond correctly to the inputs; and (ii) the capability to induce a logical failure in at least one faulty code implementation while remaining consistent with the correct execution of the reference solution. All generated test cases and test cases in $T_{\text{base}}$ are subsequently incorporated into the candidate pool $T_{\text{cand}}$. 

% \subsection{测试用例验证与筛选}
% 为确保合成的测试用例在语义上准确且能诊断出多种错误代码的同时最大化精简，我们采用了三阶段筛选流水线：输入验证，LLM指令遵循验证以及多样性筛选。
% \paragraph{输入验证}
% LLM 生成的测试用例输入往往存在“语义-执行鸿沟”，即违反了隐含的领域公理或特定问题的约束。为了消除这一鸿沟，我们参考\cite{wang2025codecontests+}的方法，为每个问题 $P_i$ 实现了一个可执行的验证器 $A_i$，并定义合法测试用例为 
% \begin{equation}
%     T_{\text{valid}} = \{t \in T_{\text{cand}} \mid A_i(t) = \text{True}\}。
% \end{equation}
% 用于过滤掉不合法的测试用例。

% \paragraph{LLM指令遵循验证}
% 经过基于验证器的输入验证剔除候选池中不满足条件的测试用例后，我们为每个候选测试进行LLM指令遵循验证。首先，通过运行参考代码对输出进行重构，以确保输入-输出的一致性。接着，通过在错误代码池上运行测试用例，以检测测试用例是否能诱发至少一个错误代码反应的逻辑错误。仅当测试用例同时满足输出正确性且能有效检测出至少一个代码缺陷时，才将其存入最终的测试用例集 $T_{\text{final}}$。

% \paragraph{多样性筛选}
% 为了在保持测试用例集精简的同时最大化不同错误模式的覆盖范围，我们根据测试用例在错误代码上的失败情况进行聚类。我们首先为$T_{\text{final}}$中每条测试用例构建二值执行向量 $V^t_i \in \{0,1\}^{|F_{\text{base}}|}$，接着利用 K-means 聚类算法将测试用例空间划分为 $K$ 个不相交的簇 $\{C_1, \dots, C_K\}$，采用的是欧几里得距离作为测试用例执行向量间差异性的指标：
% \begin{equation}
%     \min_{\{C_1, \dots, C_K\}} \sum_{k=1}^{K} \sum_{V^t_i \in C_k} \left\| V^t_i - \boldsymbol{\mu}_k \right\|_2^2
% \end{equation}
% 其中 $\boldsymbol{\mu}_k = \frac{1}{|C_k|} \sum_{V^t_j \in C_k} V^t_j$ 表示簇 $C_k$ 的质心向量。
% 对于每个生成的簇 $C_k$，我们通过识别簇中心点（Medoid）来选取代表性测试用例，即其签名最接近簇质心的用例，确保了筛选后测试集 $T_{\text{synth.}} = \{t_1, \dots, t_K\}$ 能够最大化覆盖不同错误代码对应的情况。

\subsection{Test case Validation and Selection}
\label{sec:Selection}
To ensure that the generated test cases are semantically sound and capable of diagnosing a broad spectrum of faulty code while simultaneously maximizing set parsimony, we employ a three-stage filtering pipeline comprising input verification, LLM instruction-compliance validation, and diversity-driven selection.
\paragraph{Input Validation}
LLM-generated test case inputs frequently suffer from a "semantic-execution gap," wherein they violate implicit domain axioms or problem-specific constraints. To bridge this gap, following the methodology in \cite{wang2025codecontests+}, we implement an executable validator $A_i$ for each problem $P_i$, with generation details provided in Appendix \ref{app: prompt3}. The suite of valid test cases is formally defined as:
\begin{equation}
    T_{\text{valid}} = \{t \in T_{\text{cand}} \mid A_i(t) = \text{True}\}
\end{equation}
This process prunes inputs that are syntactically correct but semantically invalid, ensuring that all subsequent evaluations are grounded in feasible execution scenarios. The input validator and its corresponding invalid test case detections are detailed in Appendix~\ref{Appendix: Input Validator}. 
\paragraph{LLM Instruction-compliance Validation}
Following the validator-based pruning of invalid test cases from the candidate pool, we subject each remaining instance to LLM instruction-compliance validation. First, the ground-truth output for each test case is re-synthesized by executing the reference code, thereby ensuring semantic alignment between inputs and outputs. Subsequently, each test case is evaluated against the faulty code pool to determine whether it can successfully trigger a logical failure in at least one faulty implementation. Only test cases that simultaneously yield correct outputs and demonstrate the capacity to expose code defects are incorporated into the final test case suite $T_{\text{final}}$. Details are in Appendix~\ref{Appendix: LLM Instruction-compliance Validation}. 
\paragraph{Diversity-Driven Selection}
To maximize the coverage of heterogeneous failure modes while maintaining test case suite parsimony, we implement a clustering mechanism based on the failure profiles of test cases against faulty code. We first construct a binary execution vector $V^t_i \in \{0,1\}^{|F_{\text{base}}|}$ for each test case in $T_{\text{final}}$, then employ the K-means algorithm~\cite{ahmed2020k} to partition the test case space into $K$ disjoint clusters $\{C_1, \dots, C_K\}$. Euclidean distance~\cite{rudin1976principles} is utilized as the dissimilarity metric to quantify differences between execution vectors:
\begin{equation}
    \min_{\{C_1, \dots, C_K\}} \sum_{k=1}^{K} \sum_{V^t_i \in C_k} \left\| V^t_i - \boldsymbol{\mu}_k \right\|_2^2
\end{equation}
where $\boldsymbol{\mu}_k = \frac{1}{|C_k|} \sum_{V^t_j \in C_k} V^t_j$ denotes the centroid of cluster $C_k$. For each resulting cluster $C_k$, a round-robin selection strategy is implemented to extract representative medoids. This strategy ensures that the final refined test case suite $T_{\text{synth.}} = \{t_1, \dots, t_K\}$ spans the maximum range of scenarios across diverse faulty codes, thereby enhancing the overall diagnostic utility. The pseudo code is presented in Appendix~\ref{sec: Algorithm}.
% \subsection{稠密奖励函数设计}

% 尽管经过了严格的双阶段数据清洗，$T_{\text{synth.}}$ 中仍可能存在残余缺陷（如未检测到的验证器漏洞或正确解法（solution）的鲁棒性不足）。为此，我们设计了一种显式建模测试集易错性的鲁棒奖励机制。
% 首先，我们将候选解 $s$ 在测试集 $T_{\text{synth.}}$ 中通过的测试用例集合定义为：
% \begin{equation}
%     \text{pass}(s, T_{\text{synth.}}) = \{ t \in T_{\text{synth.}} \mid \text{exec}(s, t) = \text{pass} \}
% \end{equation}
% 基于此，奖励函数 $r(s)$ 的计算方式如下：
% \begin{equation}
% r(s) = 
% \begin{cases} 
% 1.1 & \text{if } |\text{pass}(s,T_{\text{synth.}})| = 1 \\
% -0.1 & \text{if } |\text{pass}(s,T_{\text{synth.}})| = 0 \\
% \frac{1}{10} \cdot \frac{|\text{pass}(s,T_{\text{synth.}})|}{|T_{\text{synth.}}|} & \text{otherwise}
% \end{cases}
% \label{eq:robust_reward}
% \end{equation}

% 该奖励策略旨在通过精细化的反馈信号提升学习过程的稳定性和有效性，不仅提高了测试用例的容错性，还促进了稳定的课程学习。

\subsection{Dense Reward Function Design}
\label{sec: reward}
Despite rigorous two-stage test case pruning, residual flaws may still persist in $T_{\text{synth.}}$, such as undetected validator loopholes~\cite{liu2023your} or the insufficient robustness of reference solutions~\cite{li2022competition}. Examples are in Appendix~\ref{Appendix: Limitations of the Synthesized Test cases}. Consequently, we formulate a robust reward mechanism that explicitly accounts for the potential fallibility of the synthesized test case suite.
First, the suite of test cases successfully executed by a candidate solution $s$ within $T_{\text{synth.}}$ is defined as:
\begin{equation}
    \text{pass}(s, T_{\text{synth.}}) = \{ t \in T_{\text{synth.}} \mid \text{exec}(s, t) = \text{pass} \}
\end{equation}
The corresponding reward function $r(s)$ is formulated as follows:
\begin{equation}
r(s) = 
\begin{cases} 
1.1 & \text{if } |\text{pass}(s,T_{\text{synth.}})| = 1 \\
-0.1 & \text{if } |\text{pass}(s,T_{\text{synth.}})| = 0 \\
\frac{1}{10} \cdot \frac{|\text{pass}(s,T_{\text{synth.}})|}{|T_{\text{synth.}}|} & \text{otherwise}
\end{cases}
\label{eq:robust_reward}
\end{equation}
This reward strategy is designed to bolster the stability and efficacy of the reinforcement learning process via granular feedback signals. It not only enhances the model's resilience to potential noise within the test cases but also fosters stable curriculum learning by providing a progressive optimization trajectory.

\section{Experiment}
\begin{table*}[ht]
\centering

\begin{tabular}{lccccc} % 1(Method) + 1(Livecodebench) + 3(Codeforces) = 5列
\toprule
\multirow{2}{*}{Method} & \multicolumn{1}{c}{Livecodebench} & \multicolumn{3}{c}{Codeforces} \\
\cmidrule(lr){2-2} \cmidrule(lr){3-5}
 & Score & Score & Rating & Percentile \\
\cmidrule(lr){1-5}
Naive LLM Generation & 65.75 	& 35.41 & 83.36	& 91.70 \\
HardTests & 65.25  	& 35.65	 & 83.58 & 91.35 \\
CodeContests$^{+}$ & 65.41      & 35.56	 & 83.96 & 91.45 \\
CodeContests-O	& \underline{66.21}	& \underline{36.43}	& \underline{84.35}	& \underline{91.81} \\
\textbf{\OURS (Ours)} & \textbf{68.39}  & \textbf{38.50}	& \textbf{85.99}	& \textbf{94.67}\\
\bottomrule
\end{tabular}
\caption{Performance of baselines and \OURS{} on the LiveCodeBench and CodeForces benchmarks. The best result is bold and the second result is underline.}
\label{tab:table1}
\end{table*}

\subsection{Experimental Settings}
\paragraph{Datasets} 
% 我们采用了 CodeContests$^{+}$，这是一个综合基准数据集，汇总了来自 CodeForces、AIZU 和 AtCoder 的 11,636 个编程任务。每个问题都通过“生成器-验证器”多智能体框架生成了约 100 个测试用例。为了确保模型训练具有适当的挑战性，我们使用 Qwen3-32B 进行了基于难度的过滤。具体来说，我们评估了模型在每个问题上十次试验的性能，结果如图123所示。仅保留 pass@10 得分在 0.2 到 0.9 之间的问题参与训练，数据量约为3.3k。

% 尽管 CodeContests$^{+}$ 测试用例的规模庞大，但我们观察到其原始测试用例存在显著的同质性，这限制了基于可验证奖励的强化学习 (RLVR) 的有效性。为了克服这一问题，我们引入了 \OURDATAS{}，这是一个基于CodeContests$^{+}$构建的精细化训练集。通过使用 \OURS{} 方法，我们生成了约90个高度多样化的测试用例，能够有效地暴露潜在的逻辑漏洞。这种方法显著降低了“假阳性率”，从而增强了基于 RLVR 的训练的鲁棒性和泛化能力。
% 

\begin{figure}
    \centering
    \includegraphics[width=1\linewidth]{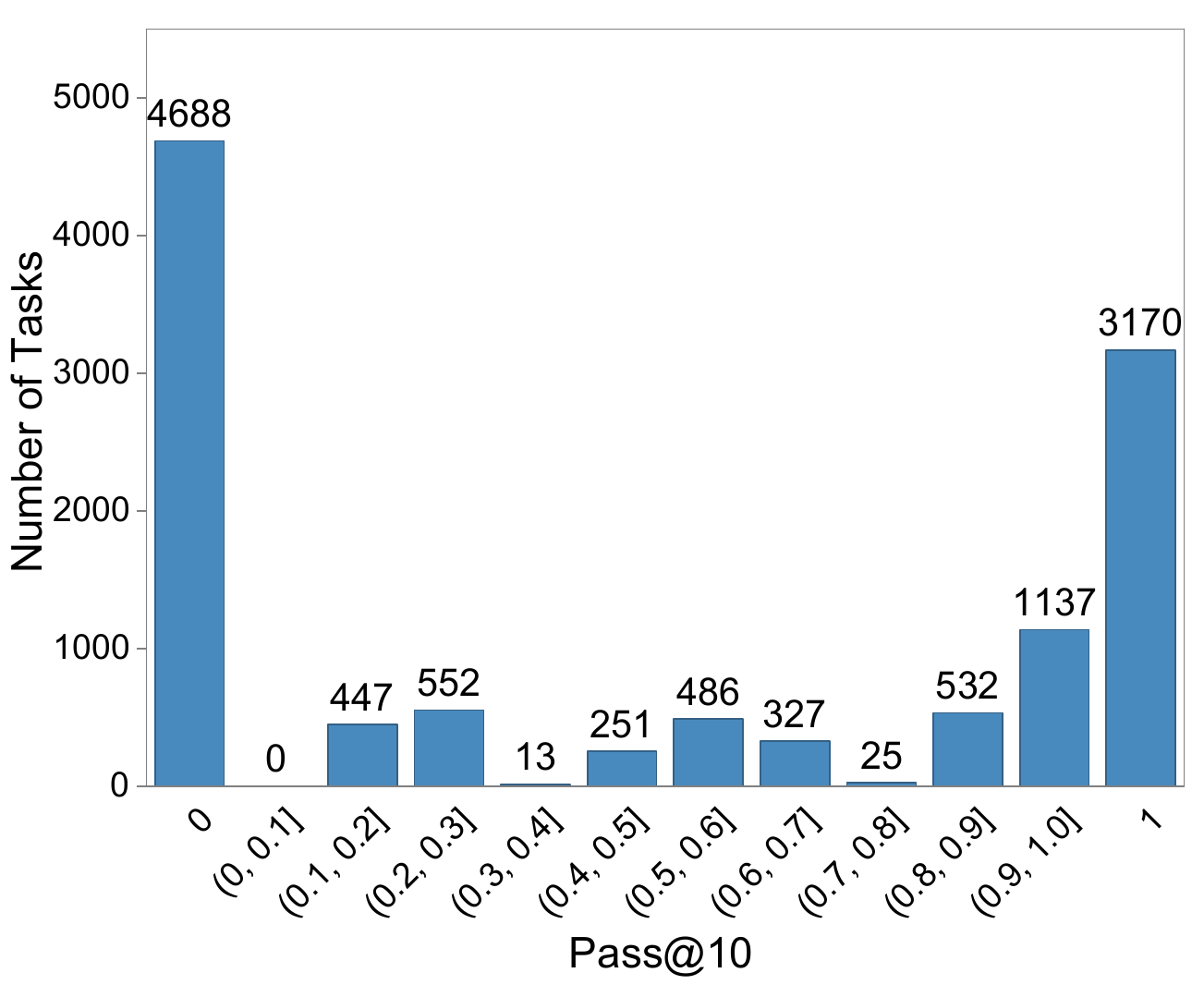}
    \caption{Correctness distribution of Qwen3-32B over ten trials across all CodeContests$^{+}$ problems, evaluated using the CodeContests$^{+}$ test case suite. The x-axis represents the distribution of pass rates over ten trials(pass@10), while the y-axis denotes the number of problems within each pass rate interval.}
    \label{fig:pass@10} 
\end{figure}

We utilize CodeContests$^{+}$~\cite{wang2025codecontests+}, a comprehensive benchmark dataset that aggregates 11,636 programming problems from CodeForces~\cite{mirzayanov2020codeforces}, AIZU~\cite{AOJ_ProgrammingChallenge, AOJ_NewSite}, and AtCoder~\cite{atcoder_website}. For each problem, approximately 100 test cases are generated via a "generator-validator" multi-agent framework. To ensure a sufficiently challenging training environment, we perform difficulty-based filtering using Qwen3-32B~\cite{yang2025qwen3}. Specifically, we assess the model's performance on ten trials for each problem, as illustrated in Figure ~\ref{fig:pass@10}. Only problems with a pass@10 score between 0.2 and 0.9 were retained, resulting in a refined training subset named $\text{CodeContests}^{+}_\text{train}$ of approximately 3.3k problems. 

Although CodeContests$^{+}$ provides a substantial volume of test cases, we observe significant homogeneity between the original instances, which constrains the efficacy of reinforcement learning from verifiable rewards (RLVR). To address the limitation, we introduce \OURS{}, a refined training set built on CodeContests$^{+}$, with approximately 200 highly diverse test cases capable of effectively uncovering latent logical defects. \OURS{} significantly reduces the false positive rate, thereby enhancing the robustness and generalization capabilities of the RLVR-based training.
\paragraph{Training Setup}
We finetune Qwen3-32B~\cite{yang2025qwen3} via GRPO~\cite{guo2025deepseek} with configuration: $\beta_{\text{clip}}=0.2$, Adam ($\eta=5\times10^{-6}$, $\beta_1=0.9$, $\beta_2=0.95$), batch size 256, and 10-step linear warmup. Answers are sampled at maximal entropy ($T=1.0$, $p_{\text{top}}=1.0$) with $n_{\text{sample}}=8$ per prompt. Input truncation at 2,048 tokens and output extension to 38,912 tokens leverage the model's 128K context window. The $\lambda_{\text{KL}}=0.0$ setting intentionally omits policy regularization, reserving KL-loss for future constrained exploration while prioritizing boundary-case discovery in high-capacity regimes. The details are in Appendix \ref{sec:Details}.
\paragraph{Baseline} 
% 我们将\OURS方法与现有的测试用例增强的方法在RLVR效果上进行了比较。这些方法包含Naive LLM Generation,HardTests,CodeContests$^{+}$。
We evaluated the efficacy of the \OURS{} in comparison with existing test case augmentation strategies within the RLVR framework. These comparative baselines encompass Naive LLM Generation~\cite{li2024large}, HardTests~\cite{he2025hardtests}, and CodeContests$^{+}$~\cite{wang2025codecontests+}, CodeContests-O~\cite{cai2026codecontests}. The introduction of baselines are in Appendix~\ref{ap: baselines}.
\paragraph{Evaluation Benchmarks} 
% 我们选取 LiveCodeBench (2024.08–2025.01) 与 CodeForces 作为评估 \OURS{} 性能的核心基准。LiveCodeBench 作为一个动态更新的基准测试，我们采用正确率 (Score) 作为评价指标。针对标准化的 CodeForces 竞赛题集，我们构建了多维度的评估体系：1) 得分 (Score)，衡量解题的准确性；2) Elo 等级分 (Rating)，通过模拟提交计算模型在排行榜上的竞技水平；3) 百分位排名 (Percentile)，用于刻画模型性能超越历史人类参赛者的比例。
We set LiveCodeBench (2024.08–2025.01) and CodeForces as the primary benchmarks to evaluate the performance of \OURS{}. For LiveCodeBench, a periodically updated benchmark, we utilize the pass rate (Score) as the evaluation metric. For the standardized CodeForces competitive problem set, we establish a multi-dimensional evaluation framework: (i) Score, which quantifies problem-solving accuracy; (ii) Rating, representing the model's competitive standing on the leaderboard determined via simulated contest participation; and (iii) Percentile, denoting the proportion of historical human contestants outperformed by the model. The evaluation settings are detailed in Appendix~\ref{sec:evaluate}.

\subsection{Main Results}
\label{mainResults}

\begin{figure}[htbp]
\centering
\begin{tikzpicture}
\begin{axis}[
    width=0.9\linewidth,
    height=5.2cm,
    ybar,
    bar width=10pt,
    ymin=0.80,
    ymax=0.94,
    ylabel={TSP},
    symbolic x coords={
        HardTests,
        CodeContests$^{+}$,
        Naive LLM Generation,
        CodeContests-O,
        RobustTests
    },
    xtick=data,
    x tick label style={
        rotate=25,
        anchor=east,
        font=\small
    },
    ylabel style={font=\small, yshift=-0.1cm},
    tick label style={font=\small},
    label style={font=\small},
    axis line style={black},
    tick style={draw=none},
    grid=none,
    enlarge x limits=0.12,
]
\addplot[
    fill={rgb,255:red,166;green,206;blue,227},
    draw={rgb,255:red,31;green,120;blue,180}
] coordinates {
    (HardTests,0.83)
    (CodeContests$^{+}$,0.84)
    (Naive LLM Generation,0.85)
    (CodeContests-O,0.88)
    (RobustTests,0.92)
};
\end{axis}

\begin{axis}[
    width=0.9\linewidth,
    height=5.2cm,
    axis y line*=right,
    axis x line=none,
    ymin=64.8,
    ymax=68.8,
    ylabel={LiveCodeBench},
    symbolic x coords={
        HardTests,
        CodeContests$^{+}$,
        Naive LLM Generation,
        CodeContests-O,
        RobustTests
    },
    xtick=data,
    ylabel style={font=\small, yshift=-0.1cm},
    tick label style={font=\small},
    label style={font=\small},
    tick style={draw=none},
    enlarge x limits=0.12,
]
\addplot[
    mark=*,
    mark size=2.0pt,
    line width=0.9pt,
    color={rgb,255:red,213;green,94;blue,0},
    mark options={fill={rgb,255:red,213;green,94;blue,0}}
] coordinates {
    (HardTests,65.25)
    (CodeContests$^{+}$,65.41)
    ({Naive LLM Generation},65.75)
    (CodeContests-O,66.21)
    (RobustTests,68.39)
};
\end{axis}
\end{tikzpicture}
\caption{Relationship between TSP values and LiveCodeBench performance across different test cases.}
\label{fig:tsp_lcb_combo}
\end{figure}
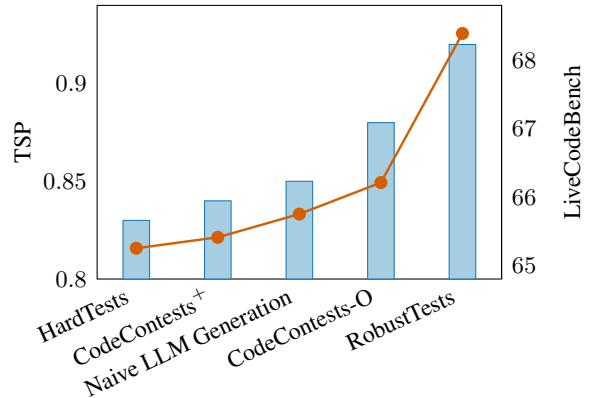

\begin{table*}[htbp]
\centering

\begin{tabular}{lccccc} % 1 (Method) + 1 (Livecodebench) + 3 (Codeforces) = 5列
\toprule
\multirow{2}{*}{Method} & Livecodebench & \multicolumn{3}{c}{Codeforces} \\
\cmidrule(lr){2-2} \cmidrule(lr){3-5}
 & Score & Score & Rating & Percentile \\
\midrule

\textbf{\OURS*(Ours)} & \textbf{67.84} & \textbf{38.47}	 & \textbf{85.16}  & \textbf{93.97} \\
w/o Test case Synthesis & 66.91	& 36.41 & 84.4	  & 92.12\\
w/o Diversity-Driven Selection & 66.82	& 36.23	& 84.12	& 91.96 \\
w/o Both Module & 65.41 & 35.56 & 83.96 & 91.45 \\
\bottomrule
\end{tabular}
\caption{Ablation study on the test case synthesis strategy of \OURS{}* on LiveCodeBench and CodeForces benchmarks. \OURS* denotes a variant of the \OURS{} framework that excludes the dense reward mechanism. We investigate the performance impact of without Test case Generation (the second row), without Diversity-Driven Selection (the third row) and without Both Module(the forth row). The best results are in bold.}
\label{tab:table2}
\end{table*}

\begin{figure}
    \centering
    \includegraphics[width=1\linewidth]{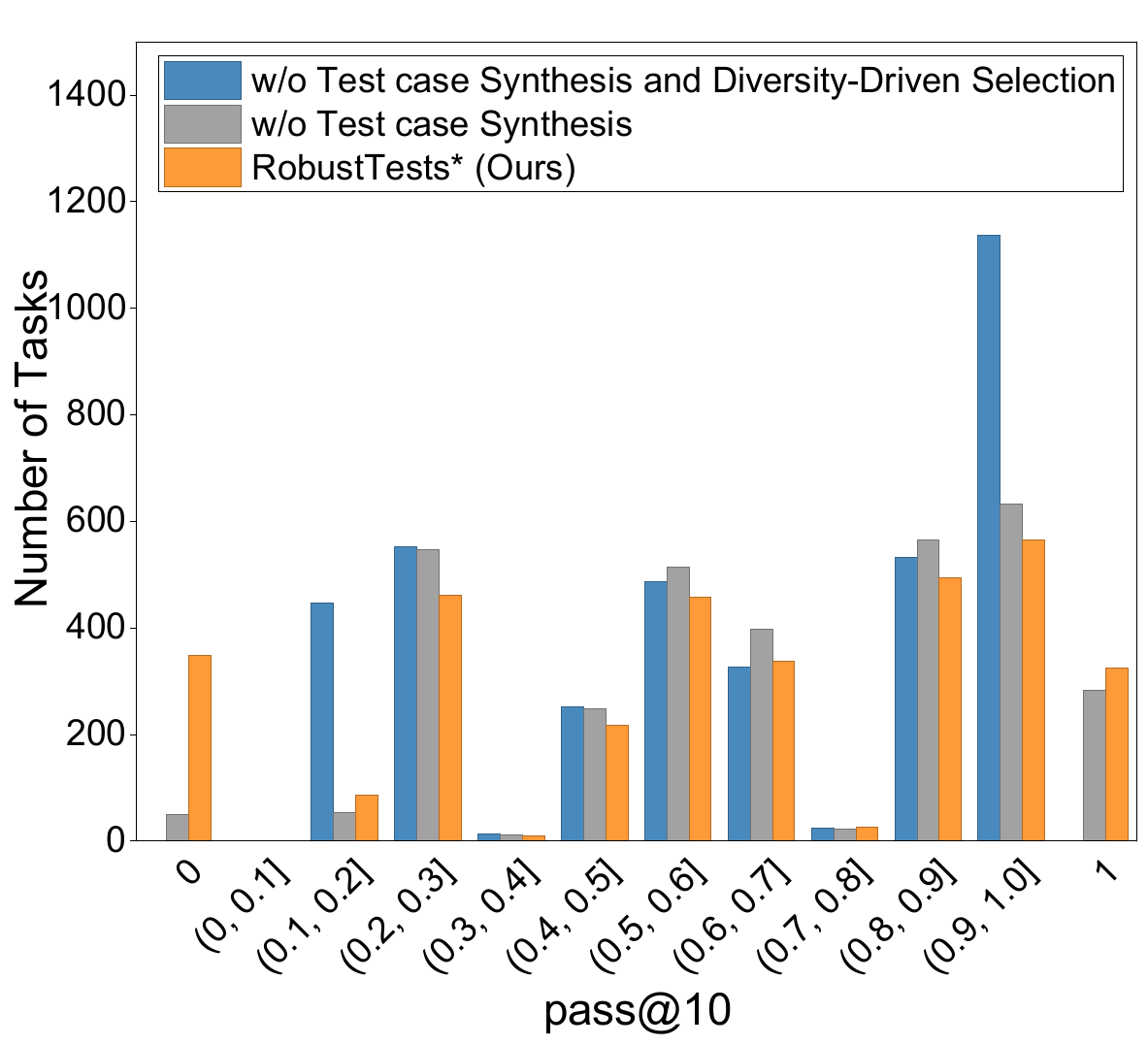}
    \caption{The pass rate distribution of Qwen3-32B over ten trials across all $\text{CodeContests}^{+}_\text{train}$ problems, evaluated against the test cases provided by \OURS{} and without Test case Generation and its configuration without Diversity-Driven Selection.}
    \label{fig:pass@10_all}
\end{figure}

\begin{figure}
    \centering
    \includegraphics[width=1\linewidth]{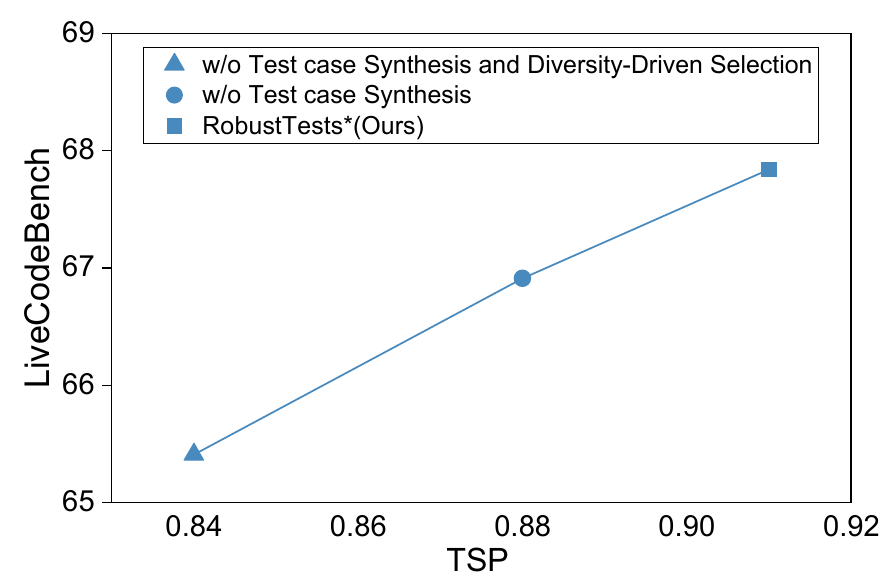}
    \caption{Correlation between the TSP values of test cases employed during training and the coding performance of LLM on the LiveCodeBench benchmark.}
    \label{fig:TSP}
\end{figure}

\begin{table*}[ht]
\centering

\begin{tabular}{lccccc} % 调整为6列：Dataset, Method, LCB-Score, CF-Score, CF-Rating, CF-Percentile
\toprule
\multirow{2}{*}{Dataset} & \multirow{2}{*}{Method} & LiveCodeBench & \multicolumn{3}{c}{Codeforces} \\
\cmidrule(lr){3-3} \cmidrule(lr){4-6}
& & Score & Score & Rating & Percentile \\
\midrule
\multirow{2}{*}{$\text{CodeContests}^{+}_\text{train}$} & Sparse Reward & 65.41 & 83.96 & 35.56 & 91.45 \\
& \textbf{Dense Reward}  & \textbf{66.30} & \textbf{84.24} & \textbf{37.74} & \textbf{92.05} \\
\midrule
\multirow{2}{*}{\OURS*} & Sparse Reward & 67.84 & 85.16 & 38.47	& 93.97 \\
& \textbf{Dense Reward}  & \textbf{68.39} & \textbf{85.99} & \textbf{38.50} & \textbf{94.67} \\
\midrule
\multirow{2}{*}{\OURS* w/o validator}
& Sparse Reward & 66.02 & 83.82 & 35.91 & 91.28 \\
& \textbf{Dense Reward} & \textbf{67.21} & \textbf{85.02} & \textbf{38.15} & \textbf{93.64} \\
\bottomrule
\end{tabular}
\caption{Ablation study on the Reward Module of \OURS{} on LiveCodeBench and CodeForces benchmarks, using both \OURS{}* and CodeContests$^{+}$ as the training datasets. The best results are in bold.}
\label{tab:table3}
\end{table*}

\begin{figure*}[htbp]
\centering
\resizebox{0.98\textwidth}{!}{
\begin{tikzpicture}
\begin{groupplot}[
    group style={
        group size=4 by 1,
        horizontal sep=1.35cm,
    },
    width=4.25cm,
    height=3.45cm,
    xlabel={Scale},
    label style={font=\small},
    tick label style={font=\footnotesize},
    axis line style={black},
    tick style={draw=none},
    xtick={0,1,2,3},
    xticklabels={0,0.05,0.10,0.20},
    enlarge x limits=0.10,
    ylabel near ticks,
    xlabel near ticks,
]

\nextgroupplot[
    title={LiveCodeBench},
    ylabel={Score},
    ymin=63.5,
    ymax=69,
]
\addplot+[mark=*, mark size=1.8pt, line width=0.85pt, color=oiBlue] coordinates {
    (0,64.05)
    (1,68.38)
    (2,68.39)
    (3,68.35)
};

\nextgroupplot[
    title={Codeforces},
    ylabel={Score},
    ymin=81.5,
    ymax=87.5,
]
\addplot+[mark=square*, mark size=1.7pt, line width=0.85pt, color=oiOrange, mark options={
        draw=oiOrange,
        fill=oiOrange
    }] coordinates {
    (0,82.11)
    (1,85.95)
    (2,85.99)
    (3,86.90)
};

\nextgroupplot[
    title={Codeforces},
    ylabel={Rating},
    ymin=34.5,
    ymax=39,
]
\addplot+[mark=triangle*, mark size=2.0pt, line width=0.85pt, color=oiGreen, mark options={
        draw=oiGreen,
        fill=oiGreen
    }] coordinates {
    (0,35.01)
    (1,38.53)
    (2,38.50)
    (3,38.48)
};

\nextgroupplot[
    title={Codeforces},
    ylabel={Percentile},
    ymin=89.5,
    ymax=95.2,
]
\addplot+[mark=diamond*, mark size=2.0pt, line width=0.85pt, color=oiGray, mark options={
        draw=oiGray,
        fill=oiGray
    }] coordinates {
    (0,90.14)
    (1,94.62)
    (2,94.67)
    (3,94.68)
};

\end{groupplot}
\end{tikzpicture}
}
\caption{Ablation study on the reward scale in the dense reward function on LiveCodeBench and Codeforces benchmarks. The scale is varied from $0$ to $0.20$ using \OURS{}$^{*}$ as the training dataset.}
\label{fig:scale_sensitivity}
\end{figure*}
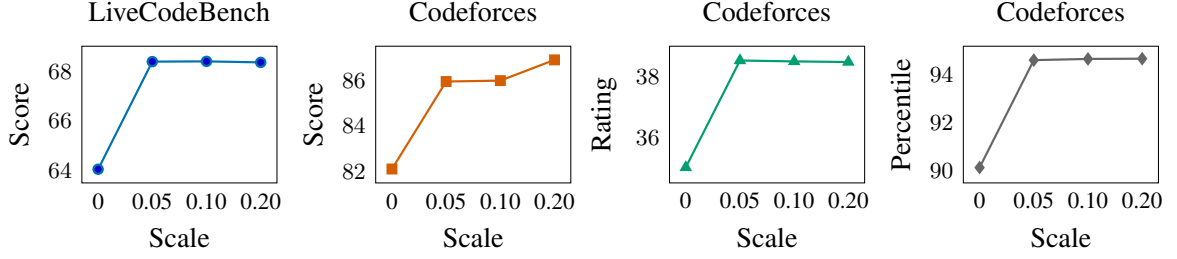

% 我们复现了HardTests和 Naive LLM Generation 方法，为CodeContests$^{+}$_train的每道题目生成了一些测试用例。为了公平起见，所有方法均将测试用例的个数限制在40个左右。在HardTests，Naive LLM Generation和CodeContests$^{+}$方法下，我们采用的是0-1稀疏奖励对Qwen3-32B进行RL。在\OURS{}方法下，接着采用的是稠密奖励对Qwen3-32B进行RL。我们将不同的方法进行了对比，结果表明，三个baseline在LiveCodeBench和CodeForces的表现基本持平，我们的方法在这两个榜单上比baseline均要高3%左右。这种性能的提升一方面源于训练集中测试用例能更有效地检测出代码存在的逻辑问题，减少"假阳性"的情况，另一方面源于采用了基于通过率的阶梯式稠密奖励函数的设计，该设计不仅能在测试用例不可避免出现错误时给予部分正确的奖励信号，从而减少"假阴性"的情况，还实现了课程学习的引导效果。
\paragraph{Performance on LiveCodeBench and CodeForces} We reimplement HardTests and Naive LLM Generation methods to synthesize test cases for each problem in $\text{CodeContests}^{+}_\text{train}$. To ensure a fair comparison, the number of test cases per problem is capped at approximately 40 across all methods. For HardTests, Naive LLM Generation, and CodeContests$^{+}$, CodeContests-O baselines, we conduct reinforcement learning on Qwen3-32B using binary (0-1) sparse rewards. In contrast, the \OURS{} method employs dense rewards for reinforcement learning on the same model. The results are shown in Table~\ref{tab:table1}, revealing that while the three baselines exhibited comparable performance on LiveCodeBench and CodeForces, \OURS{} outperform them by approximately absolute 3\% on both benchmarks. This performance gain is twofold: First, the high-quality test cases in our training set more effectively detect logical discrepancies in answers, thereby reducing false positives. Second, the stepwise dense reward function provides intermediate feedback even when hallucinations of test cases are unavoidable. The design not only mitigates false negatives but also facilitates a curriculum learning effect that guides the model toward absolute correctness.
\paragraph{Analysis of Test case Diversity} 
To better understand why RobustTests improves downstream training, we analyze the diversity of different test cases. Drawing on~\cite{jaeger2000landscape}, we introduce Test case Space Polarization (TSP), a low-cost metric that quantifies the diagnostic coverage of a test case suite across diverse faulty codes:
\begin{equation}
\begin{aligned}
TSP = \frac{1}{M} \cdot  \sum_{s \in S} \left(-\frac{n_s}{N} \cdot \log_2 \frac{n_s}{N}  \right)
\end{aligned}
\end{equation}
Here, $M$ denotes the number of faulty codes in $F_{\text{base}}$, $N$ is the total number of test cases in $T_{\text{synth.}}$, and $S$ represents the set of test cases in $T_{\text{synth.}}$. The term $n_s$ indicates the frequency of a specific test case $s$ in $T_{\text{synth.}}$. As shown in Figure~\ref{fig:tsp_lcb_combo}, RobustTests achieves the highest TSP value among all compared test cases and obtains the best LiveCodeBench performance. In contrast, methods with lower TSP values, such as HardTests and CodeContests$^{+}$, lead to weaker downstream performance. This trend suggests that TSP is positively correlated with the training utility of test cases and can serve as a low-cost proxy for estimating dataset potential before expensive RL training.
\subsection{Ablation Study}

%在本节中，我们进行了一项广泛的消融研究，以评估测试用例生成策略中每个组件对整体性能的贡献。此外，我们引入了一种新的指标test case Space Polarization（TSP），通过测量测试用例对各种错误代码实现的诊断覆盖率来量化其多样性，并评估其对模型效果的影响。最后，我们研究了训练阶段所采用的不同奖励函数对模型性能的影响。
In this section, we conduct an extensive ablation study to assess the individual contribution of each component within the test case synthesis strategy to the aggregate performance. Furthermore, we introduce a novel metric, Test case Space Polarization (TSP), designed to quantify test case diversity by measuring their diagnostic coverage across faulty codes, and evaluate its subsequent influence on model performance. Finally, we investigate the impact of various reward functions used during the training stage on the efficacy of the model.
\paragraph{Test case Synthesis Module Ablation}
% 我们首先研究了Diversity-Driven Selection和Test case Synthesis阶段对整体效果的贡献，结果如Table \ref{tab:table2}所示。在所有实验中，我们控制了测试用例的数量为40左右，以及训练方式均采用0-1稀疏奖励训练模型。可以看出，当两者结合使用时，Qwen3-32B性能最佳，这证明了它们的互补性：Test case Synthesis为Test case suit中添加了能够有效区分正确代码与错误代码的高质量测试用例，Diversity-Driven Selection能筛选出最大化覆盖不同错误代码的测试用例。
We first investigated the individual contributions of the Diversity-Driven Selection and Test case Synthesis stages to model performance, with results summarized in Table~\ref{tab:table2}. In all experimental configurations, we fix the test case budget at approximately 40 and trained all models using binary (0-1) sparse rewards. The findings indicate that Both Test Case Synthesis and Diversity-Driven Selection yield consistent gains of approximately 1.5\% on LiveCodeBench when applied independently and the integrated approach yields the best performance for Qwen3-32B, demonstrating a clear complementarity between the two stages: Test case Synthesis introduces high-quality test cases into the suite that are capable of effectively distinguishing correct codes from faulty code, while Diversity-Driven Selection prunes the suite to retain test cases that maximize coverage across diverse faulty code.

We further analyze the role of test case diversity in the ablation study using the TSP metric introduced in Section~\ref{mainResults}. As shown in Figure~\ref{fig:TSP}, under identical training configurations, training Qwen3-32B with test cases that have higher TSP values leads to stronger LiveCodeBench performance. This indicates that higher diagnostic diversity improves the quality of reward signals by reducing the likelihood that faulty solutions are mistakenly accepted as correct.

Furthermore, we study the relationship between TSP and the pass rate of model outputs under different test cases. Figure~\ref{fig:pass@10_all} shows the pass-rate distribution over ten trials on 3.3k problems from $\text{CodeContests}^{+}_{\text{train}}$ using Qwen3-32B. After removing Test Case Synthesis and Diversity-Driven Selection from \OURS{}, TSP decreases and the distribution shifts rightward, indicating that lower-diversity test cases are less effective at detecting faulty solutions and therefore introduce more false positives in reward assignment. In contrast, higher-TSP test cases provide stricter diagnostic signals and better distinguish correct solutions from faulty ones.

Interestingly, as TSP increases, the pass@10 of some problems reaches 1. Further analysis suggests that this is caused by remaining spurious synthetic test cases, which can falsely reject semantically correct solutions and create persistent false negatives. This motivates both validator filtering and dense rewards: the former removes invalid test cases, while the latter lets the model learn from partial execution feedback instead of noisy binary rewards.

\paragraph{Reward Module Ablation}
% Table ~\ref{tab:table3} demonstrates the efficacy gains of dense rewards over sparse rewards, using test cases in \OURS{} and in CodeContests$^{+}$. 
% It is evident that, regardless of whether test cases in \OURS{} or CodeContests$^{+}$ are employed for the training set, the integration of the dense reward function yields an improvement of approximately 0.8 percentage points on the LiveCodeBench benchmark. This underscores the efficacy and generalizability of our proposed stepwise dense reward function. Given that synthetic test cases inevitably harbor instances that violate problem constraints, inducing false negatives, the dense reward mechanism enables the model to derive informative signals from partially correct feedback. This capacity for continuous learning from granular signals ultimately bolsters the code generation proficiency of the model.
Table~\ref{tab:table3} shows that dense rewards outperform sparse rewards across different training test cases, including \OURS{}, CodeContests$^{+}$, and RobustTests$^{*}$ without validator filtering. Dense rewards consistently improve LiveCodeBench performance with both \OURS{} and CodeContests$^{+}$, demonstrating the effectiveness and generalizability of our stepwise dense reward function. Under the no-validator RobustTests$^{*}$ setting, dense rewards yield even larger gains on both LiveCodeBench and Codeforces, indicating stronger robustness to noisy or invalid test cases.

To explain this result, we audit the validator filtering pipeline on approximately 3.3k problems from CodeContests$^{+}_{\text{train}}$. Each problem contains about 200 raw generated test cases, around 30\% of which are rejected as invalid, with no reference-execution failures observed. However, manual post-hoc inspection shows that about 10\% of the accepted test cases remain invalid, causing false-negative judgments for roughly 10\% of all test cases. These findings suggest that validator filtering removes many invalid test cases but cannot fully eliminate test noise. Thus, when validator filtering is weakened or removed, dense rewards provide a more robust training signal by leveraging partial execution feedback and reducing the impact of false-negative sparse rewards.

We further examine reward-scale sensitivity in the dense reward function. As shown in Figure~\ref{fig:scale_sensitivity}, setting the scale to zero causes a clear performance drop, confirming the necessity of this component. Performance remains stable as the scale varies from $0.05$ to $0.20$, suggesting that this hyperparameter has limited influence once enabled.

\section{Conclusion}
% 我们提出了一种包含错误代码驱动的测试用例构造与稠密奖励机制的设计的方法\OURS{}来提供更鲁棒性的代码RL训练，并据此构造了数据集\OURDATAS{}。我们所提出的测试用例构造方法通过提升测试用例对不同错误代码的诊断覆盖率，从而减少了模型误将正确代码判断为错误代码的情况，即"假阳性率"。我们设计的稠密奖励机制在合成测试用例具有不可避免的错误，从而会产生“假阴性”的情况下，模型可以从部分正确信号中持续学习，同时提供了类课程学习的效果。通过将错误代码驱动的测试用例构造和稠密奖励机制的设计，\OURS{}为大模型的代码生成能力的提升开辟了一条有前景的路径。
We propose ~\OURS{}, integrates faulty-code-driven test case synthesis and a stepwise dense reward mechanism to establish a robust RL framework for code generation, significantly enhancing the diagnostic utility of the augmented CodeContests$^{+}$ dataset. By leveraging "near-correct" faulty codes to expand diagnostic coverage, the framework effectively minimizes false positives, while the stepwise dense reward mitigates false negatives by enabling the model to learn from partially correct signals through a curriculum learning paradigm. Beyond code generation, these principles offer broad applicability: the synthesis strategy can be adapted for automated mutation testing in software engineering, and the dense reward mechanism is particularly suited for experimental planning and conclusion analysis in autonomous scientific discovery. By bridging these capabilities, ~\OURS{} provides a versatile trajectory for enhancing reasoning and planning within complex scientific and engineering contexts.
\section*{Limitations}
Although \OURS{} successfully strengthens the coding generation abilities of LLMs, certain limitations persist that we aim to mitigate in subsequent work:
\begin{itemize}
    \item Dependency on solutions: The proposed approach is primarily applicable to programming tasks with available ground-truth solutions; consequently, its utility is constrained in real-world scenarios where reference implementations are absent.
    \item Expansion of domain generalization: While the proposed framework is evaluated on competitive programming benchmarks such as LiveCodeBench and CodeForces, its generalizability to a broader range of software development tasks remains to be fully explored.
\end{itemize}

\section*{Ethical Considerations}
The Use of AI Assistants We employed Gemini-3 to assist us in polishing our paper and coding.

\section*{Acknowledgments}
This work was supported by Ant Group Research Intern Program.

\bibliography{custom}

@article{el2025competitive,
  title={Competitive programming with large reasoning models},
  author={El-Kishky, Ahmed and Wei, Alexander and Saraiva, Andre and Minaiev, Borys and Selsam, Daniel and Dohan, David and Song, Francis and Lightman, Hunter and Clavera, Ignasi and Pachocki, Jakub and others},
  journal={arXiv preprint arXiv:2502.06807},
  year={2025}
}

@article{guo2025deepseek,
  title={Deepseek-r1: Incentivizing reasoning capability in llms via reinforcement learning},
  author={Guo, Daya and Yang, Dejian and Zhang, Haowei and Song, Junxiao and Wang, Peiyi and Zhu, Qihao and Xu, Runxin and Zhang, Ruoyu and Ma, Shirong and Bi, Xiao and others},
  journal={arXiv preprint arXiv:2501.12948},
  year={2025}
}

@article{he2025hardtests,
  title={Hardtests: Synthesizing high-quality test cases for llm coding},
  author={He, Zhongmou and Choi, Yee Man and Zhang, Kexun and Ji, Jiabao and Zhou, Junting and Xu, Dejia and Bercovich, Ivan and Zhang, Aidan and Li, Lei},
  journal={arXiv preprint arXiv:2505.24098},
  year={2025}
}

@article{wang2025codecontests+,
  title={Codecontests+: High-quality test case generation for competitive programming},
  author={Wang, Zihan and Liu, Siyao and Sun, Yang and Li, Hongyan and Shen, Kai},
  journal={arXiv preprint arXiv:2506.05817},
  year={2025}
}

@inproceedings{zeng2025acecoder,
  title={Acecoder: Acing coder rl via automated test-case synthesis},
  author={Zeng, Huaye and Jiang, Dongfu and Wang, Haozhe and Nie, Ping and Chen, Xiaotong and Chen, Wenhu},
  booktitle={Proceedings of the 63rd Annual Meeting of the Association for Computational Linguistics (Volume 1: Long Papers)},
  pages={12023--12040},
  year={2025}
}

@article{jiang2026survey,
  title={A survey on large language models for code generation},
  author={Jiang, Juyong and Wang, Fan and Shen, Jiasi and Kim, Sungju and Kim, Sunghun},
  journal={ACM Transactions on Software Engineering and Methodology},
  volume={35},
  number={2},
  pages={1--72},
  year={2026},
  publisher={ACM New York, NY}
}

@article{team2025every,
  title={Every step evolves: Scaling reinforcement learning for trillion-scale thinking model},
  author={Team, Ling and Shen, Anqi and Li, Baihui and Hu, Bin and Jing, Bin and Chen, Cai and Huang, Chao and Zhang, Chao and Yang, Chaokun and Lin, Cheng and others},
  journal={arXiv preprint arXiv:2510.18855},
  year={2025}
}

@inproceedings{xu2025kodcode,
  title={Kodcode: A diverse, challenging, and verifiable synthetic dataset for coding},
  author={Xu, Zhangchen and Liu, Yang and Yin, Yueqin and Zhou, Mingyuan and Poovendran, Radha},
  booktitle={Findings of the Association for Computational Linguistics: ACL 2025},
  pages={6980--7008},
  year={2025}
}

@article{kwa2024catastrophic,
  title={Catastrophic Goodhart: regularizing RLHF with KL divergence does not mitigate heavy-tailed reward misspecification},
  author={Kwa, Thomas and Thomas, Drake and Garriga-Alonso, Adri{\`a}},
  journal={Advances in Neural Information Processing Systems},
  volume={37},
  pages={14608--14633},
  year={2024}
}

@article{yang2025qwen3,
  title={Qwen3 technical report},
  author={Yang, An and Li, Anfeng and Yang, Baosong and Zhang, Beichen and Hui, Binyuan and Zheng, Bo and Yu, Bowen and Gao, Chang and Huang, Chengen and Lv, Chenxu and others},
  journal={arXiv preprint arXiv:2505.09388},
  year={2025}
}

@article{jain2024livecodebench,
  title={Livecodebench: Holistic and contamination free evaluation of large language models for code},
  author={Jain, Naman and Han, King and Gu, Alex and Li, Wen-Ding and Yan, Fanjia and Zhang, Tianjun and Wang, Sida and Solar-Lezama, Armando and Sen, Koushik and Stoica, Ion},
  journal={arXiv preprint arXiv:2403.07974},
  year={2024}
}

@article{lin2025learning,
  title={Learning to solve and verify: A self-play framework for code and test generation},
  author={Lin, Zi and Shen, Sheng and Shang, Jingbo and Weston, Jason and Nie, Yixin},
  journal={arXiv preprint arXiv:2502.14948},
  year={2025}
}

@inproceedings{ma2025dynamic,
  title={Dynamic scaling of unit tests for code reward modeling},
  author={Ma, Zeyao and Zhang, Xiaokang and Zhang, Jing and Yu, Jifan and Luo, Sijia and Tang, Jie},
  booktitle={Proceedings of the 63rd Annual Meeting of the Association for Computational Linguistics (Volume 1: Long Papers)},
  pages={6917--6935},
  year={2025}
}

@article{le2022coderl,
  title={Coderl: Mastering code generation through pretrained models and deep reinforcement learning},
  author={Le, Hung and Wang, Yue and Gotmare, Akhilesh Deepak and Savarese, Silvio and Hoi, Steven Chu Hong},
  journal={Advances in Neural Information Processing Systems},
  volume={35},
  pages={21314--21328},
  year={2022}
}

@article{liu2023your,
  title={Is your code generated by chatgpt really correct? rigorous evaluation of large language models for code generation},
  author={Liu, Jiawei and Xia, Chunqiu Steven and Wang, Yuyao and Zhang, Lingming},
  journal={Advances in neural information processing systems},
  volume={36},
  pages={21558--21572},
  year={2023}
}

@article{gunjal2025rubrics,
  title={Rubrics as rewards: Reinforcement learning beyond verifiable domains},
  author={Gunjal, Anisha and Wang, Anthony and Lau, Elaine and Nath, Vaskar and He, Yunzhong and Liu, Bing and Hendryx, Sean},
  journal={arXiv preprint arXiv:2507.17746},
  year={2025}
}

@article{olausson2023self,
  title={Is self-repair a silver bullet for code generation?},
  author={Olausson, Theo X and Inala, Jeevana Priya and Wang, Chenglong and Gao, Jianfeng and Solar-Lezama, Armando},
  journal={arXiv preprint arXiv:2306.09896},
  year={2023}
}

@article{casper2023open,
  title={Open problems and fundamental limitations of reinforcement learning from human feedback},
  author={Casper, Stephen and Davies, Xander and Shi, Claudia and Gilbert, Thomas Krendl and Scheurer, J{\'e}r{\'e}my and Rando, Javier and Freedman, Rachel and Korbak, Tomasz and Lindner, David and Freire, Pedro and others},
  journal={arXiv preprint arXiv:2307.15217},
  year={2023}
}

@article{yang2025swe,
  title={Swe-smith: Scaling data for software engineering agents},
  author={Yang, John and Lieret, Kilian and Jimenez, Carlos E and Wettig, Alexander and Khandpur, Kabir and Zhang, Yanzhe and Hui, Binyuan and Press, Ofir and Schmidt, Ludwig and Yang, Diyi},
  journal={arXiv preprint arXiv:2504.21798},
  year={2025}
}

@article{ouyang2022training,
  title={Training language models to follow instructions with human feedback},
  author={Ouyang, Long and Wu, Jeffrey and Jiang, Xu and Almeida, Diogo and Wainwright, Carroll and Mishkin, Pamela and Zhang, Chong and Agarwal, Sandhini and Slama, Katarina and Ray, Alex and others},
  journal={Advances in neural information processing systems},
  volume={35},
  pages={27730--27744},
  year={2022}
}

@article{ahmed2020k,
  title={The k-means algorithm: A comprehensive survey and performance evaluation},
  author={Ahmed, Mohiuddin and Seraj, Raihan and Islam, Syed Mohammed Shamsul},
  journal={Electronics},
  volume={9},
  number={8},
  pages={1295},
  year={2020},
  publisher={MDPI}
}

@book{rudin1976principles,
  title={Principles of Mathematical Analysis},
  author={Rudin, Walter},
  year={1976},
  publisher={McGraw-Hill},
  address={New York}
}

@inproceedings{mu2024ddprompt,
  title={Ddprompt: Differential diversity prompting in large language models},
  author={Mu, Lin and Zhang, Wenhao and Zhang, Yiwen and Jin, Peiquan},
  booktitle={Proceedings of the 62nd Annual Meeting of the Association for Computational Linguistics (Volume 2: Short Papers)},
  pages={168--174},
  year={2024}
}

@article{mirzayanov2020codeforces,
  title={Codeforces as an educational platform for learning programming in digitalization},
  author={Mirzayanov, Mike and Pavlova, Oksana and Mavrin, Pavel and Melnikov, Roman and Plotnikov, Andrew and Parfenov, Vladimir and Stankevich, Andrew},
  journal={Olympiads in Informatics},
  volume={14},
  number={133-142},
  pages={14},
  year={2020}
}

@misc{AOJ_ProgrammingChallenge,
  key = {AOJ Programming Challenge},
  title = {Aizu online judge: Programming challenge},
  howpublished = {\url{http://judge.u-aizu.ac.jp/onlinejudge/}},
  note = {Accessed: 23 Apr. 2018},
  year = {2018}
}

@misc{AOJ_NewSite,
  key = {AOJ New Site},
  title = {Aizu online judge (new site)},
  howpublished = {\url{http://onlinejudge.u-aizu.ac.jp/home}},
  note = {Accessed: 23-Apr-2018},
  year = {2018}
}

@misc{atcoder_website,
  author = {{AtCoder Inc.}},
  title = {AtCoder: Programming Contest Website},
  year = {2012},
  howpublished = {\url{https://atcoder.jp/}},
}

@article{li2024large,
  title={Large language models as test case generators: Performance evaluation and enhancement},
  author={Li, Kefan and Yuan, Yuan},
  journal={arXiv preprint arXiv:2404.13340},
  year={2024}
}

@article{jaeger2000landscape,
  title={Landscape division, splitting index, and effective mesh size: new measures of landscape fragmentation},
  author={Jaeger, Jochen AG},
  journal={Landscape ecology},
  volume={15},
  number={2},
  pages={115--130},
  year={2000},
  publisher={Springer}
}

@article{austin2021program,
  title={Program synthesis with large language models},
  author={Austin, Jacob and Odena, Augustus and Nye, Maxwell and Bosma, Maarten and Michalewski, Henryk and Dohan, David and Jiang, Ellen and Cai, Carrie and Terry, Michael and Le, Quoc and others},
  journal={arXiv preprint arXiv:2108.07732},
  year={2021}
}

@article{chen2021evaluating,
  title={Evaluating large language models trained on code},
  author={Chen, Mark and Tworek, Jerry and Jun, Heewoo and Yuan, Qiming and Pinto, Henrique Ponde De Oliveira and Kaplan, Jared and Edwards, Harri and Burda, Yuri and Joseph, Nicholas and Brockman, Greg and others},
  journal={arXiv preprint arXiv:2107.03374},
  year={2021}
}

@article{li2022competition,
  title={Competition-level code generation with alphacode},
  author={Li, Yujia and Choi, David and Chung, Junyoung and Kushman, Nate and Schrittwieser, Julian and Leblond, R{\'e}mi and Eccles, Tom and Keeling, James and Gimeno, Felix and Dal Lago, Agustin and others},
  journal={Science},
  volume={378},
  number={6624},
  pages={1092--1097},
  year={2022},
  publisher={American Association for the Advancement of Science}
}

@article{schulman2017proximal,
  title={Proximal policy optimization algorithms},
  author={Schulman, John and Wolski, Filip and Dhariwal, Prafulla and Radford, Alec and Klimov, Oleg},
  journal={arXiv preprint arXiv:1707.06347},
  year={2017}
}

@article{hou2024large,
  title={Large language models for software engineering: A systematic literature review},
  author={Hou, Xinyi and Zhao, Yanjie and Liu, Yue and Yang, Zhou and Wang, Kailong and Li, Li and Luo, Xiapu and Lo, David and Grundy, John and Wang, Haoyu},
  journal={ACM Transactions on Software Engineering and Methodology},
  volume={33},
  number={8},
  pages={1--79},
  year={2024},
  publisher={ACM New York, NY}
}

@inproceedings{dou2024stepcoder,
  title={Stepcoder: improving code generation with reinforcement learning from compiler feedback},
  author={Dou, Shihan and Liu, Yan and Jia, Haoxiang and Zhou, Enyu and Xiong, Limao and Shan, Junjie and Huang, Caishuang and Wang, Xiao and Fan, Xiaoran and Xi, Zhiheng and others},
  booktitle={Proceedings of the 62nd Annual Meeting of the Association for Computational Linguistics (Volume 1: Long Papers)},
  pages={4571--4585},
  year={2024}
}

@article{shojaee2023execution,
  title={Execution-based code generation using deep reinforcement learning},
  author={Shojaee, Parshin and Jain, Aneesh and Tipirneni, Sindhu and Reddy, Chandan K},
  journal={arXiv preprint arXiv:2301.13816},
  year={2023}
}

@article{peng2023yarn,
  title={Yarn: Efficient context window extension of large language models},
  author={Peng, Bowen and Quesnelle, Jeffrey and Fan, Honglu and Shippole, Enrico},
  journal={arXiv preprint arXiv:2309.00071},
  year={2023}
}

@article{team2025kimi,
  title={Kimi k2: Open agentic intelligence},
  author={Team, Kimi and Bai, Yifan and Bao, Yiping and Chen, Guanduo and Chen, Jiahao and Chen, Ningxin and Chen, Ruijue and Chen, Yanru and Chen, Yuankun and Chen, Yutian and others},
  journal={arXiv preprint arXiv:2507.20534},
  year={2025}
}

@article{cai2026codecontests,
  title={CodeContests-O: Powering LLMs via Feedback-Driven Iterative Test Case Generation},
  author={Cai, Jianfeng and Zhu, Jinhua and Sun, Ruopei and Zhao, Kangwen and Xue, Dongyun and Feng, Mingxiao and Zhou, Wengang and Li, Houqiang},
  journal={arXiv preprint arXiv:2601.13682},
  year={2026}
}

\newpage
\appendix

\section{Training Settings}
\label{sec:Details}
% 我们在32张H200上展开实验，使用的基模为Qwen3-32B。实验的具体参数如Table \ref{tab: details}所示。
We conduct our experiments on 32 NVIDIA H200 GPUs, employing Qwen3-32B as the base model. The detailed experimental parameters are summarized in Table~\ref{tab: details}.
\begin{table}[h]
\centering

\begin{tabular}{lc}
\toprule
\textbf{Parameter} & \textbf{Value} \\
\midrule
Optimizer & AdamW \\
Learning Rate ($\eta$) & $5 \times 10^{-6}$ \\
Adam $\beta_1, \beta_2$ & $0.9, 0.95$ \\
Weight Decay & $0.01$ \\
Gradient Clipping & $1.0$ \\
Global Batch Size & $256$ \\
Warmup Steps & $10$ \\
Learning Rate Scheduler & Linear \\
$n_\text{sample}$ & $8$ \\
Clipping Range ($\beta_{\text{clip}}$) & $0.2$ \\
KL Coefficient ($\lambda_{\text{KL}}$) & $0.0$ \\
Temperature ($T$) & $1.0$ \\
Top-$p$ & $1.0$ \\
Max Input Tokens & $2,048$ \\
Max Output Tokens & $38,912$ \\
Numerical Precision & BF16 \\
\bottomrule
\end{tabular}
\caption{Hyperparameters for Experiment}
\label{tab: details}
\end{table}

\section{Evaluation Settings}
\label{sec:evaluate}
Following the evaluation protocol established in ~\cite{team2025every}, we assess a standardized pipeline to ensure a rigorous and fair comparison. For benchmarks including LiveCodeBench(2024.08–2025.01) and CodeForces, evaluations are conducted using a 128K context window. For the Qwen3-32B model with limited native context length, we leverage YaRN~\cite{peng2023yarn} for context extension and strictly adhere to the official hyperparameters specified for its open-weight release.

\section{Introduction of Baselines}
\label{ap: baselines}
\paragraph{Naive LLM Generation} Naive LLM Generation~\cite{li2024large} is an approach that directly generates test cases. Specifically, it first prompts the LLM to directly synthesize a test case suite that conforms to the problem specifications, and then validates and revises the generated test cases with the reference solution.
\paragraph{HardTests} HardTests~\cite{he2025hardtests} is an approach that generates test cases via dedicated generator programs. Specifically, the LLM is first prompted to produce generator programs that can automatically synthesize test case inputs, and the corresponding outputs are then obtained using the reference solution.
\paragraph{CodeContests$^{+}$} CodeContests$^{+}$~\cite{wang2025codecontests+} is an approach that generates test cases via a "generator-validator" multi-agent framework. It first produces test case inputs using programs generated by the LLM, and then validates these inputs with an input validator also generated by the LLM. The corresponding outputs are subsequently obtained using the reference solution.
\paragraph{CodeContests-O} CodeContests-O~\cite{cai2026codecontests} is a Feedback-Driven Iterative Framework that transforms test case synthesis from open-loop generation into a closed-loop process by utilizing execution feedback from both correct and incorrect solutions to refine test cases for high fidelity and discriminability.

\section{Case Study}
%在这一章节中，我们将以CodeContests$^{+}$中一道题目为例，详细介绍测试用例合成的各个组件是如何运作的以及必要性,和存在不合法测试用例的原因。
In this section, we utilize a representative problem from CodeContests$^{+}$ as a case study to provide a detailed exposition of the operational mechanics and the underlying necessity of each component in the test case synthesis process, while further elucidating the root causes behind the generation of invalid test cases.
\subsection{Problem}
% 我们选取了CodeContests$^{+}$数据集中id为p03520的题目。
We identify problem p03520 from the CodeContests$^{+}$ dataset. The detailed problem statement is presented as follows:

\begin{lstlisting}[style=NormalStyle]
Snuke found a record of a tree with N vertices in ancient ruins. The findings are as follows:
* The vertices of the tree were numbered 1,2,...,N, and the edges were numbered 1,2,...,N-1.
* Edge i connected Vertex a_i and b_i.
* The length of each edge was an integer between 1 and 10^{18} (inclusive).
* The sum of the shortest distances from Vertex i to Vertex 1,...,N was s_i.
From the information above, restore the length of each edge. The input guarantees that it is possible to determine the lengths of the edges consistently with the record. Furthermore, it can be proved that the length of each edge is uniquely determined in such a case.
Constraints
* 2 \leq N \leq 10^{5}
* 1 \leq a_i,b_i \leq N
* 1 \leq s_i \leq 10^{18}
* The given graph is a tree.
* All input values are integers.
* It is possible to consistently restore the lengths of the edges.
* In the restored graph, the length of each edge is an integer between 1 and 10^{18} (inclusive).
(*@\textbf{Input}@*)
Input is given from Standard Input in the following format:
N
a_1 b_1
:
a_{N-1} b_{N-1}
s_1 s_2 ... s_{N}
(*@\textbf{Output}@*)
Print N-1 lines. The i-th line must contain the length of Edge i.
\end{lstlisting}
\subsection{Original Test cases}
\label{sec:Originalt}
%这道题目原有的测试用例数量为177个，即$|T_\text{base} = 177|$。示例如下：
The base test suite for this problem comprises 177 test cases, denoted as $|T_{\text{base}}| = 177$. An illustrative example is provided below:
\begin{lstlisting}[style=NormalStyle]
(*@\textbf{Input}@*)
5
1 2
1 3
1 4
1 5
10 13 16 19 22
(*@\textbf{Output}@*)
1
2
3
4
\end{lstlisting}
\subsection{Faulty Codes and Generated Test cases}
\label{appendix: Faulty Codes and Generated Test cases}
% 在章节\ref{sec:Synthesis}中，我们根据题目生成了一些错误代码，语言为python3，以生成能区分正确代码和错误代码的测试用例。该题目生成的符合条件的错误代码$|F_\text{base} = 12|$，示例如下:
As described in Section \ref{sec:Synthesis}, we generate a collection of faulty codes to guide the synthesis of discriminative test cases. The total number of qualified faulty implementations for the problem is $|F_{\text{base}}| = 12$, with an example provided as follows:
\begin{lstlisting}[style=NormalStyle]
import sys
input=sys.stdin.readline
def int_lst():
    return list(map(int,input().split()))
N=int(input())
adj=[[]for _ in range(N)]
edges=[]
for idx in range(N-1):
    a,b=int_lst()
    a-=1;b-=1
    edges.append((a,b,idx))
    adj[a].append((b,idx))
    adj[b].append((a,idx))
s=int_lst()
sz=[0]*N
def dfs(v,p):
    cnt=1
    for u,idx in adj[v]:
        if u!=p:
            cnt+=dfs(u,v)
    sz[v]=cnt
    return cnt
dfs(0,-1)
ans=[0]*(N-1)
for a,b,i in edges:
    if sz[a]<sz[b]:
        ans[i]=(s[b]-s[a])//(sz[a]-sz[b])
    else:
        ans[i]=(s[a]-s[b])//(sz[b]-sz[a])
for val in ans:print(abs(val))
\end{lstlisting}
% 接着，我们利用生成的错误代码引导LLM生成了一些能区分正确代码和错误代码的测试用例，共有75个，示例如下：
Subsequently, we leverage the synthesized faulty code to guide the LLM in generating test cases capable of discriminating between correct and faulty implementations. The process yields a total of 75 test cases, a representative example of which is provided below:
\begin{lstlisting}[style=NormalStyle]
(*@\textbf{Input}@*)
6
1 2
2 3
3 4
4 5
5 6
15 12 9 9 12 15
(*@\textbf{Output}@*)
3
2
1
2
3
\end{lstlisting}
% 该测试用例的输出是错误的，因此后续我们需要通过solution来修正测试用例的输出。由于$T_\text{cand}$包含了$T_\text{base}$中的测试用例，因此|T_\text{cand}| = 252。
The outputs of these test cases are initially placeholder values and require calibration against the reference solution. Since $T_{\text{cand}}$ is constructed by merging the synthesized test cases with $T_{\text{base}}$, the resulting candidate set size is $|T_{\text{cand}}| = 252$.
\subsection{Input Validator}
\label{Appendix: Input Validator}
% 我们参考~\cite{wang2025codecontests+}的方法为该题生成了输入验证器，如下所示:
Drawing on the approach of ~\cite{wang2025codecontests+}, an input validator is implemented for this problem as follows:
\begin{lstlisting}[style=NormalStyle]
#include "testlib.h"
#include <bits/stdc++.h>
using namespace std;

const int MAXN = 100000;
const long long MAXSI = 1000000000000000000LL;

int parent[MAXN + 5];

int find(int x) {
    if (parent[x] != x)
        parent[x] = find(parent[x]);
    return parent[x];
}
void unite(int x, int y) {
    x = find(x);
    y = find(y);
    if (x != y)
        parent[x] = y;
}

int main(int argc, char* argv[]) {
    registerValidation(argc, argv);
    int n = inf.readInt(2, 100000);
    inf.readEoln();
    for (int i = 1; i <= n; ++i)
        parent[i] = i;
    set<pair<int, int>> edges;
    for(int i = 0; i < n - 1; i++) {
        int a = inf.readInt(1, n);
        inf.readSpace();
        int b = inf.readInt(1, n);
        inf.readEoln();
        ensuref(a != b, "Self-loop detected at edge %d", i+1);
        int u = min(a, b);
        int v = max(a, b);
        ensuref(edges.count({u, v}) == 0, "Multiple edges between %d and %d", u, v);
        edges.insert({u, v});
        ensuref(find(a) != find(b), "Cycle detected while adding edge between %d and %d", a, b);
        unite(a, b);
    }
    // Check connectedness
    int root = find(1);
    for(int i = 2; i <= n; i++) {
        ensuref(find(i) == root, "Graph is not connected, node %d is in different component", i);
    }
    // Read s_i
    vector<long long> s = inf.readLongs(n, 1, MAXSI);
    inf.readEoln();
    inf.readEof();
    return 0;
}
\end{lstlisting}
% 该验证器检验出2个不满足题意的测试用例，其中一个如下：
The validator flags two test cases for failing to satisfy the problem constraints, one of which is illustrated as follows.
\begin{lstlisting}[style=NormalStyle]
(*@\textbf{Input}@*)
8
1 2
2 3
3 4
4 5
5 6
6 7
7 8
28 24 20 16 12 8 4 0
(*@\textbf{Output}@*)
4
4
4
4
4
4
4
\end{lstlisting}
% 该测试用例的输入包含一个等于 0 的整数，但这违反了题目设定的范围限制 $[1,10^18]$
This test case includes an integer value of 0, thereby violating the problem's defined input range of $[1, 10^{18}]$.
\subsection{LLM Instruction-compliance Validation}
\label{Appendix: LLM Instruction-compliance Validation}
% 我们对LLM生成的测试用例是否要求进行了验证。一方面，我们利用Solution对生成的测试用例的输出进行了修正,~\ref{sec:Originalt}中的测试用例修正后如下：
To ensure the validity of the LLM-synthesized test cases, we implement a verification and refinement process. First, the test case outputs are calibrated using the reference solution; the updated version of the case mentioned in Appendix~\ref{sec:Originalt} is illustrated as follows:
\begin{lstlisting}[style=NormalStyle]
(*@\textbf{Input}@*)
6
1 2
2 3
3 4
4 5
5 6
15 12 9 9 12 15
(*@\textbf{Output}@*)
3
1
3
1
3
\end{lstlisting}
On the other hand, we prune non-discriminative test cases that are passed by the faulty implementations. In this instance, two test cases were filtered out, resulting in a final test suite of $|T_{\text{final}}| = 248$ after the two-stage refinement process.
% 另一方面，我们过滤掉了所有错误代码能通过的测试用例。对于本题，我们过滤掉了2个测试用例。经过两阶段过滤后，$|T_\text{final} = 248|$。
\subsection{Diversity-Driven Selection}
\label{Appendix: Diversity-Driven Selection}
% 我们按照实验要求筛选了40个具有代表性的测试用例，使得$|T_\text{synth.} = 40|$。筛选过程如~\ref{sec: Algorithm}所示。
In accordance with our experimental requirements, we select 40 representative test cases such that $|T_{\text{synth.}}| = 40$. The formal selection process is outlined in Appendix~\ref{sec: Algorithm}.
\subsection{Limitations of the Synthesized Test cases}
\label{Appendix: Limitations of the Synthesized Test cases}
% 输入验证器没有检查给定的 s 数组是否能够对应到一组合法的整数边长。导致$T_\text{synth.}$中的部分测试用例被误判为合法的测试用例，该测试用例如下：
Due to the absence of a check for the existence of valid integer side lengths corresponding to array $s$, several test cases in $T_{\text{synth.}}$ are falsely accepted by the input validator. An example of the hallucination test cases is illustrated as follows:
\begin{lstlisting}[style=NormalStyle]
(*@\textbf{Input}@*)
8
1 2
2 3
3 4
4 5
5 6
6 7
7 8
32 28 22 16 16 22 28 32
(*@\textbf{Output}@*)
0
1
3
0
3
1
0
\end{lstlisting}
% Qwen3-32B的正确回答无法通过该测试用例，被误判为错误，从而导致了“假阴性”的现象。因此，需要基于步长的稠密奖励函数来进行不完美测试用例的容错。

\begin{algorithm*}[ht] % [t] 表示固定在页面顶部 (Top)
\caption{Diversity-Driven Test case Selection}
\label{alg:cds}
\begin{algorithmic}[1]
\REQUIRE Initial test case suite $T_{\text{final}}$, target cluster count $K$, maximum budget $M$;
\ENSURE Refined test case suite $T_{\text{synth.}}$;
\FOR{each test case $t_i \in T_{\text{final}}$}
    \STATE Construct binary execution vector $V^t_i \in \{0,1\}^{|F_{\text{base}}|}$ based on failure profiles;
\ENDFOR
\STATE $X \leftarrow \{V^t_1, V^t_2, \dots, V^t_n\}$;

\STATE $X \leftarrow \text{Standardize}(X)$;

\STATE $K' \leftarrow \min(K, |T_{\text{final}}|)$;
\STATE $\{\mathcal{L}, \boldsymbol{\mu}\} \leftarrow \text{K-Means}(X, \text{n\_clusters}=K')$; 
\STATE Partition $T_{\text{final}}$ into clusters $\{C_1, C_2, \dots, C_{K'}\}$ based on $\mathcal{L}$;
\FOR{each cluster $k \in \{1, \dots, K'\}$}
    \FOR{each test case $t_{k,i} \in C_k$}
        \STATE Compute dissimilarity to centroid: $\delta_{k,i} = \| V^t_{k,i} - \boldsymbol{\mu}_k \|_2$;
    \ENDFOR
    \STATE Sort $C_k$ in ascending order of $\delta_{k,i}$ (closest to centroid first);
\ENDFOR
\STATE $T_{\text{synth.}} \leftarrow \emptyset$, $j \leftarrow 0$;
\WHILE{$|T_{\text{synth.}}| < M$ \AND $|T_{\text{synth.}}| < |T_{\text{final}}|$}
    \FOR{$k = 1$ \TO $K'$}
        \IF{$j < |C_k|$}
            \STATE $T_{\text{synth.}} \leftarrow T_{\text{synth.}} \cup \{C_k[j]\}$;
            \IF{$|T_{\text{synth.}}| = M$}
                \STATE \textbf{break};
            \ENDIF
        \ENDIF
    \ENDFOR
    \STATE $j \leftarrow j + 1$;
\ENDWHILE
\RETURN $T_{\text{synth.}}$
\end{algorithmic}
\end{algorithm*}

The rejection of correct answers from Qwen3-32B by this test case leads to false negatives. To address this, we employ a stepwise dense reward function to enhance the model's robustness against imperfect test cases.

\section{Algorithm definition}
\label{sec: Algorithm}
% 在章节\ref{sec:Selection}中，我们通过聚类的方式对测试用例进行了多样性筛选.具体的算法定义如Algorithm~\ref{alg:cds}所示。
In Section \ref{sec:Selection}, we employ clustering techniques to filter test cases for enhanced diversity. A pseudo code of it is provided as follows in Algorithm~\ref{alg:cds}. 
% First, we construct binary execution vectors $V^t_i$ to characterize the failure profiles of test cases. Then, K-means clustering is employed to partition the test case space into $K$ disjoint clusters $\{C_1, \dots, C_K\}$. Within each cluster, test cases are ranked based on their Euclidean distance to the centroid $\boldsymbol{\mu}_k$. Finally, a round-robin selection strategy is implemented to extract representative medoids, forming a parsimonious suite $T_{\text{synth.}}$ that maximizes the coverage of heterogeneous failure modes. We provide a pseudo code of it as follows so that the readers can easily understand the whole learning procedure. The detailed procedure of the algorithm is presented in Algorithm~\ref{alg:cds}.   

\section{Implementation Details}
This section details the specific prompt templates employed in three stages described in Section~\ref{sec:method}, including Faulty Code Generation, Faulty-Code-Driven Test Case Generation and Input Validation Generation, with Kimi-k2~\cite{team2025kimi} serving as the underlying foundation model.
\subsection{Faulty Code Generation}
\label{app: prompt1}
For each problem, we generate faulty code through a multi-prompt approach. As illustrated in Figures ~\ref{fig:fc1}, ~\ref{fig: fc2} and ~\ref{fig: fc3}, we incorporate three distinct prompts, each of which is executed for three independent sampling passes. This results in a total of nine model invocations to ensure high intra-class diversity among the generated faulty implementations.
\subsection{Faulty-Code-Driven Test Case Generation}
\label{app: prompt2}
For each problem, we generate a set of discriminative test cases capable of differentiating between correct and faulty codes. The prompt is illustrated in Figure ~\ref{fig: fc4}.
\subsection{Input Validator Generation}
\label{app: prompt3}
For each problem, we construct input validators designed to ensure that the synthesized test case inputs strictly adhere to the problem specifications. The prompt is illustrated in Figure ~\ref{fig: fc5}.
\begin{figure*}

\begin{tcolorbox}[
    colback=white,
    colframe=black!75,
    title=\textbf{Prompt: Generate faulty codes that fail on edge cases}, 
    fonttitle=\bfseries,
    boxrule=0.8pt,
    arc=2mm,
    left=6pt, right=6pt, top=6pt, bottom=6pt,
    boxsep=2pt
]
% \small

System message: \\
Your input fields are: \\
quesition: \$title + \$description \\
solution: \$solution-code \\

Your output fields are:\\
1. `reasoning` (str): \\
2. `faulty codes` (str): Generate a **diverse and comprehensive** set of faulty codes to distinguish high-quality test cases.\\

All interactions will be structured in the following way, with the appropriate values filled in.\\
{}[[ \#\# question \#\# ]]\\
{}[[ \#\# solution \#\# ]]\\
{}[[ \#\# reasoning \#\# ]] \\
{}[[ \#\# faulty codes \#\# ]] \\
{}[[ \#\# completed \#\# ]]\\
In adhering to this structure, your objective is: Given the fields `quesition` and `solution`, produce the fields `faulty code` as many as possible, don't produce the repeat `faulty code`. \\

faulty codes should satisfy: \\
(i) The code language is Python 3.\\
(ii) The code should pass a subset of the given test cases (e.g., simple/typical cases) but fail on edge cases.\\
(iii) The faults should be subtle (e.g., integer overflow, boundary condition omission, incorrect loop termination, type confusion, unhandled exceptional cases), avoiding obvious syntax errors.\\
(iv) The code structure should remain reasonable, avoiding "obviously wrong" patterns (e.g., deliberate division by zero).\\

The format of faulty codes should satisfy: please provide the faulty codes in JSON-like format(a list of dictionaries): \\
```json\\
{}[{"code": <string>}]\\
```\\
Ensure the faulty code match the types and structure expected for the question. Do not include any additional text or explanations, just the JSON-like object. \\

User message: \\
{}[[ \#\# question \#\# ]] \$title + \$description \\

Respond with the corresponding output fields, starting with the field `{}[[ \#\# reasoning \#\# ]]`, then
`{}[[ \#\# faulty codes \#\# ]]`, and then ending with the marker for `{}[[ \#\# completed \#\# ]]`.
\end{tcolorbox}
\caption{One of prompts on Faulty Code Generation Stage}
\label{fig:fc1}
\end{figure*}

\begin{figure*}

\begin{tcolorbox}[
    colback=white,
    colframe=black!75,
    title=\textbf{Prompt: Generate faulty codes that fail on hidden logical paths}, 
    fonttitle=\bfseries,
    boxrule=0.8pt,
    arc=2mm,
    left=6pt, right=6pt, top=6pt, bottom=6pt,
    boxsep=2pt
]

System message: \\
Your input fields are: \\
quesition: \$title + \$description \\
solution: \$solution-code \\

Your output fields are:\\
1. `reasoning` (str): \\
2. `faulty codes` (str): Generate a **diverse and comprehensive** set of faulty codes to distinguish high-quality test cases.\\

All interactions will be structured in the following way, with the appropriate values filled in.\\
{}[[ \#\# question \#\# ]]\\
{}[[ \#\# solution \#\# ]]\\
{}[[ \#\# reasoning \#\# ]] \\
{}[[ \#\# faulty codes \#\# ]] \\
{}[[ \#\# completed \#\# ]]\\
In adhering to this structure, your objective is: Given the fields `quesition` and `solution`, produce the fields `faulty code` as many as possible, don't produce the repeat `faulty code`. \\

faulty codes should satisfy: \\
(i) The code language is Python 3.\\
(ii) The code should pass a subset of the given test cases (e.g., simple/typical cases) but fail on hidden logical paths.\\
(iii) The faults should be subtle (e.g., integer overflow, boundary condition omission, incorrect loop termination, type confusion, unhandled exceptional cases), avoiding obvious syntax errors.\\
(iv) The code structure should remain reasonable, avoiding "obviously wrong" patterns (e.g., deliberate division by zero).\\

The format of faulty codes should satisfy: please provide the faulty codes in JSON-like format(a list of dictionaries): \\
```json\\
{}[{"code": <string>}]\\
```\\
Ensure the faulty code match the types and structure expected for the question. Do not include any additional text or explanations, just the JSON-like object. \\

User message: \\
{}[[ \#\# question \#\# ]] \$title + \$description \\

Respond with the corresponding output fields, starting with the field `{}[[ \#\# reasoning \#\# ]]`, then
`{}[[ \#\# faulty codes \#\# ]]`, and then ending with the marker for `{}[[ \#\# completed \#\# ]]`.
\end{tcolorbox}
\caption{One of prompts on Faulty Code Generation Stage}
\label{fig: fc2}
\end{figure*}

\begin{figure*}

\begin{tcolorbox}[
    colback=white,
    colframe=black!75,
    title=\textbf{Prompt: Generate faulty codes that fail on special inputs}, 
    fonttitle=\bfseries,
    boxrule=0.8pt,
    arc=2mm,
    left=6pt, right=6pt, top=6pt, bottom=6pt,
    boxsep=2pt
]

System message: \\
Your input fields are: \\
quesition: \$title + \$description \\
solution: \$solution-code \\

Your output fields are:\\
1. `reasoning` (str): \\
2. `faulty codes` (str): Generate a **diverse and comprehensive** set of faulty codes to distinguish high-quality test cases.\\

All interactions will be structured in the following way, with the appropriate values filled in.\\
{}[[ \#\# question \#\# ]]\\
{}[[ \#\# solution \#\# ]]\\
{}[[ \#\# reasoning \#\# ]] \\
{}[[ \#\# faulty codes \#\# ]] \\
{}[[ \#\# completed \#\# ]]\\
In adhering to this structure, your objective is: Given the fields `quesition` and `solution`, produce the fields `faulty code` as many as possible, don't produce the repeat `faulty code`. \\

faulty codes should satisfy: \\
(i) The code language is Python 3.\\
(ii) The code should pass a subset of the given test cases (e.g., simple/typical cases) but fail on special inputs.\\
(iii) The faults should be subtle (e.g., integer overflow, boundary condition omission, incorrect loop termination, type confusion, unhandled exceptional cases), avoiding obvious syntax errors.\\
(iv) The code structure should remain reasonable, avoiding "obviously wrong" patterns (e.g., deliberate division by zero).\\

The format of faulty codes should satisfy: please provide the faulty codes in JSON-like format(a list of dictionaries): \\
```json\\
{}[{"code": <string>}]\\
```\\
Ensure the faulty code match the types and structure expected for the question. Do not include any additional text or explanations, just the JSON-like object. \\

User message: \\
{}[[ \#\# question \#\# ]] \$title + \$description \\

Respond with the corresponding output fields, starting with the field `{}[[ \#\# reasoning \#\# ]]`, then
`{}[[ \#\# faulty codes \#\# ]]`, and then ending with the marker for `{}[[ \#\# completed \#\# ]]`.
\end{tcolorbox}
\caption{One of prompts on Faulty Code Generation Stage}
\label{fig: fc3}
\end{figure*}

\begin{figure*}

\begin{tcolorbox}[
    colback=white,
    colframe=black!75,
    title=\textbf{Prompt: Generate test cases that can differentiate between correct code and faulty code}, 
    fonttitle=\bfseries,
    boxrule=0.8pt,
    arc=2mm,
    left=6pt, right=6pt, top=6pt, bottom=6pt,
    boxsep=2pt
]

System message: \\
Your input fields are:\\
quesition: \$title + \$description\\
solution: \$solution\_code\\
faulty\_code: \$faulty\_code\\

Your output fields are:\\
1. `reasoning` (str): \\
2. `test cases` (str): Generate a diverse and comprehensive set of test cases that can pass the solution but fail on the faulty code.\\

All interactions will be structured in the following way, with the appropriate values filled in.\\
{}[[ \#\# question \#\# {}]]\\
{}[[ \#\# solution \#\# {}]]\\
{}[[ \#\# faulty\_code \#\# {}]]\\
{}[[ \#\# reasoning \#\# {}]] \\
{}[[ \#\# test cases \#\# {}]] \\
{}[[ \#\# completed \#\# {}]]\\
In adhering to this structure, your objective is: Given the fields `question`, `solution` and `faulty\_code`, produce the fields `test cases` as many as possible. The `test cases` must satisfy the requirement that they can pass the solution but fail on the faulty code. This is a mandatory requirement. Please do not generate any `test cases` that do not meet this criterion.\\

The format of test case should satisfy: please provide the input and output in JSON-like format(a list of dictionaries): \\
```json\\
{}[{{"input": <string>, "output": <string>}}{}]\\
```\\
Ensure the input and output match the types and structure expected for the problem. Do not include any additional text or explanations, just the JSON-like object.\\

User message:\\
{}[[ \#\# question \#\# {}]] \$title + \$description\\

{}[[ \#\# solution \#\# {}]] \$solution\_code\\

{}[[ \#\# faulty\_code \#\# {}]] \$faulty\_code\\

Respond with the corresponding output fields, starting with the field `{}[[ \#\# reasoning \#\# {}]]`, then `{}[[ \#\# test cases \#\# {}]]`, and then ending with the marker for `{}[[ \#\# completed \#\# {}]]`. \\
\end{tcolorbox}
\caption{Prompt on Faulty-Code-Driven Test Case Generation Stage}
\label{fig: fc4}
\end{figure*}

\begin{figure*}

\begin{tcolorbox}[
    colback=white,
    colframe=black!75,
    title=\textbf{Prompt: Generate the input validator}, 
    fonttitle=\bfseries,
    boxrule=0.8pt,
    arc=2mm,
    left=6pt, right=6pt, top=6pt, bottom=6pt,
    boxsep=2pt
]

System message: \\
Your input fields are:

question: \$title + \$description. \\

Your output fields are:\\
reasoning (str): Explain how to implement a robust validator using \texttt{testlib.h}, including the validation of basic types, ranges, whitespace, and complex geometric constraints (e.g., simple polygon check, no duplicate points).\\
validator\_cpp (str): The complete, production-ready C++ source code for the input validator.
All interactions must be structured exactly as follows: \\
{}[[ \#\# question \#\# {}]]\\
{}[[ \#\# reasoning \#\# {}]]\\
{}[[ \#\# validator\_cpp \#\# {}]]\\
{}[[ \#\# completed \#\# {}]]\\

Objective: \\
Given the \texttt{question}, produce the \texttt{validator\_cpp} using the \texttt{testlib.h} library. The validator must satisfy the following mandatory requirements:

Strict Formatting: Use \texttt{registerValidation(argc, argv)}, \texttt{inf.readInt()}, \texttt{inf.readSpace()}, \texttt{inf.readEoln()}, and \texttt{inf.readEof()}.
Constraint Verification: Ensure all variables (e.g., $N$, coordinates) are within the specified bounds. \\

The format of test case should satisfy: please provide in JSON-like format(a list of dictionaries): \\
```json\\
{}[{{"validator\_cpp": <string>}}{}]\\
```\\
Ensure the validator\_cpp match the types and structure expected for the problem. Do not include any additional text or explanations, just the JSON-like object.\\

User Message: \\
{}[[ \#\# question \#\# {}]] \\
\$title + \$description\\

Instructions for Response: \\
Respond with the corresponding output fields, starting with {}[[ \#\# reasoning \#\# {}]], followed by {}[[ \#\# validator\_cpp \#\# {}]], and ending with the marker {}[[ \#\# completed \#\# {}]].
\end{tcolorbox}
\caption{Prompt on Input Validator Generation Stage}
\label{fig: fc5}
\end{figure*}

\end{document}